\documentclass{article}

\usepackage[table]{xcolor}
\usepackage{array}
\usepackage{booktabs}
\usepackage[preprint]{corl_2026} 

\newcommand{\methodname}{\textit{GLIDE}}

\newcommand{\glidetag}{\text{\textsc{GLIDE}}}
\newcommand{\detag}{\text{\textsc{DE}}}
\newcommand{\Gglide}[1]{\ensuremath{G_{\glidetag}^{(#1)}}}
\newcommand{\Ginit}{\Gglide{0}}
\newcommand{\Giter}{\Gglide{k}}
\newcommand{\Gnext}{\Gglide{k+1}}
\newcommand{\Gopt}{\ensuremath{G_{\glidetag}^{*}}}
\newcommand{\Gde}{\ensuremath{G_{\detag}}}
\newcommand{\Dround}[1]{\ensuremath{\mathcal{D}^{(#1)}}}
\newcommand{\Diter}{\Dround{k}}
\newcommand{\Dopt}{\ensuremath{\mathcal{D}^{*}}}
\newcommand{\Dpool}{\ensuremath{\mathcal{D}_{\mathrm{pool}}}}
\newcommand{\Gsetglide}{\ensuremath{\mathcal{G}_{\glidetag}}}
\newcommand{\Gsetde}{\ensuremath{\mathcal{G}_{\detag}}}
\newcommand{\teleopcode}{\ensuremath{T}}
\newcommand{\pihum}{\ensuremath{\pi_{\mathrm{hum}}}}
\newcommand{\Dsucc}{\ensuremath{\mathcal{D}_{\mathrm{pool}}^{\mathrm{succ}}}}
\newcommand{\Dimpf}{\ensuremath{\mathcal{D}_{\mathrm{pool}}^{\mathrm{impf}}}}
\newcommand{\Dtrainsucc}{\ensuremath{\mathcal{D}_{\mathrm{train}}^{\mathrm{succ}}}}
\newcommand{\Dtrainmix}{\ensuremath{\mathcal{D}_{\mathrm{train}}^{\mathrm{mix}}}}

\usepackage{graphicx} %
\usepackage{wrapfig} %
\usepackage{amsmath,amssymb} %
\usepackage{twemojis} %
\usepackage{listings} %
\usepackage{longtable} %

\newcommand{\tomatoemoji}{\raisebox{-0.06ex}{\twemoji[height=0.72em]{tomato}}}
\newcommand{\markeremoji}{\raisebox{-0.06ex}{\twemoji[height=0.72em]{crayon}}}
\newcommand{\wineemoji}{\raisebox{-0.06ex}{\twemoji[height=0.72em]{wine_glass}}}
\definecolor{datasuccessgreen}{rgb}{0.25,0.78,0.36}
\definecolor{datafailgray}{gray}{0.78}
\definecolor{tasktomatobg}{RGB}{222,240,202}
\definecolor{taskmarkerbg}{RGB}{179,231,250}
\definecolor{taskwinebg}{RGB}{255,236,179}
\newcommand{\datablock}[1]{%
  \begingroup
  \textcolor{#1}{\rule{0.72em}{0.52em}}%
  \endgroup
}
\newcommand{\successdata}{\datablock{datasuccessgreen}}
\newcommand{\mixeddata}{\successdata\hspace{0.12em}\datablock{datafailgray}}
\newcommand{\vanilladata}{\datablock{black}}
\newcommand{\guardyes}{\textcolor{datasuccessgreen}{$\checkmark$}}
\newcommand{\guardno}{\textcolor{red!70!black}{$\times$}}
\lstdefinestyle{promptblock}{
  basicstyle=\ttfamily\footnotesize,
  aboveskip=0.25em,
  belowskip=0.85em,
  breakautoindent=false,
  breakatwhitespace=true,
  breakindent=0pt,
  breaklines=true,
  columns=fullflexible,
  frame=single,
  framerule=0.4pt,
  framexleftmargin=0.45em,
  framexrightmargin=0.45em,
  framextopmargin=0.25em,
  framexbottommargin=0.25em,
  keepspaces=false,
  linewidth=\linewidth,
  showstringspaces=false,
  xleftmargin=0pt,
  xrightmargin=0pt
}

\newcommand{\figref}[1]{Fig.~\ref{#1}}
\newcommand{\secref}[1]{Sec.~\ref{#1}}
\newcommand{\tabref}[1]{Tab.~\ref{#1}}
\newcounter{algorithm}
\newcommand{\algref}[1]{Alg.~\ref{#1}}
\newcounter{algorithmline}[algorithm]
\renewcommand{\thealgorithmline}{\arabic{algorithmline}}
\newcommand{\algline}[2]{\refstepcounter{algorithmline}\label{#1}\thealgorithmline: & #2 \\}

\definecolor{algcommentblue}{RGB}{0,92,175}
\newcommand{\algcomment}[1]{\unskip\hfill\mbox{\textcolor{algcommentblue}{$\triangleright$~\textit{#1}}}}
\renewcommand{\eqref}[1]{Eq.~\ref{#1}}

\title{Learning Beyond What Humans Can Demonstrate}

\author{
  Yuchen Song, Aditya Mittal, Unnat Jain \\
  University of California, Irvine
}

\begin{document}
\maketitle


\begin{abstract}
Behavior cloning for robot manipulation relies on
expert demonstrations. However, for tasks that require dynamic
stability, precise contact timing, or dexterous coordination,
human operators may find it hard or even impossible to collect data. We study this
infeasible-demonstration regime and propose \methodname{}: Guardrails for Learning from Infeasible Demonstrations Efficiently, a framework that
infers task-specific failure modes and converts them into executable
guardrails for data collection and policy deployment. Given a task description
and the conditioning teleoperation code, \methodname{} writes guardrails that use system
states to filter teleoperation and
policy commands, constrain failure-prone actions, and iteratively improve from trajectory
feedback. Across three tasks, \methodname{} discovers emergent guardrails that
go beyond domain-expert hardcoded ones, improving data collection over naive
VR teleoperation and domain-expert hardcoded guardrails. After refinement,
\methodname{} raises data-collection success from 0--10\% to
70--90\% across the three tasks. During policy execution, mixed-data guarded
policies reach 70\%, 60\%, and 60\% success on \emph{Tomato plate transfer},
\emph{Marker handover \& stand}, and \emph{Wine serving} tasks. These results show that
\methodname{} can support policy learning when direct
demonstrations are infeasible. Project website:
\url{http://guardrail-policy.github.io/}.
\end{abstract}



\section{Introduction}
\label{sec:introduction}

\begin{wrapfigure}[16]{r}{0.48\linewidth}
	\vspace{-1em}
	\centering
	\includegraphics[width=\linewidth]{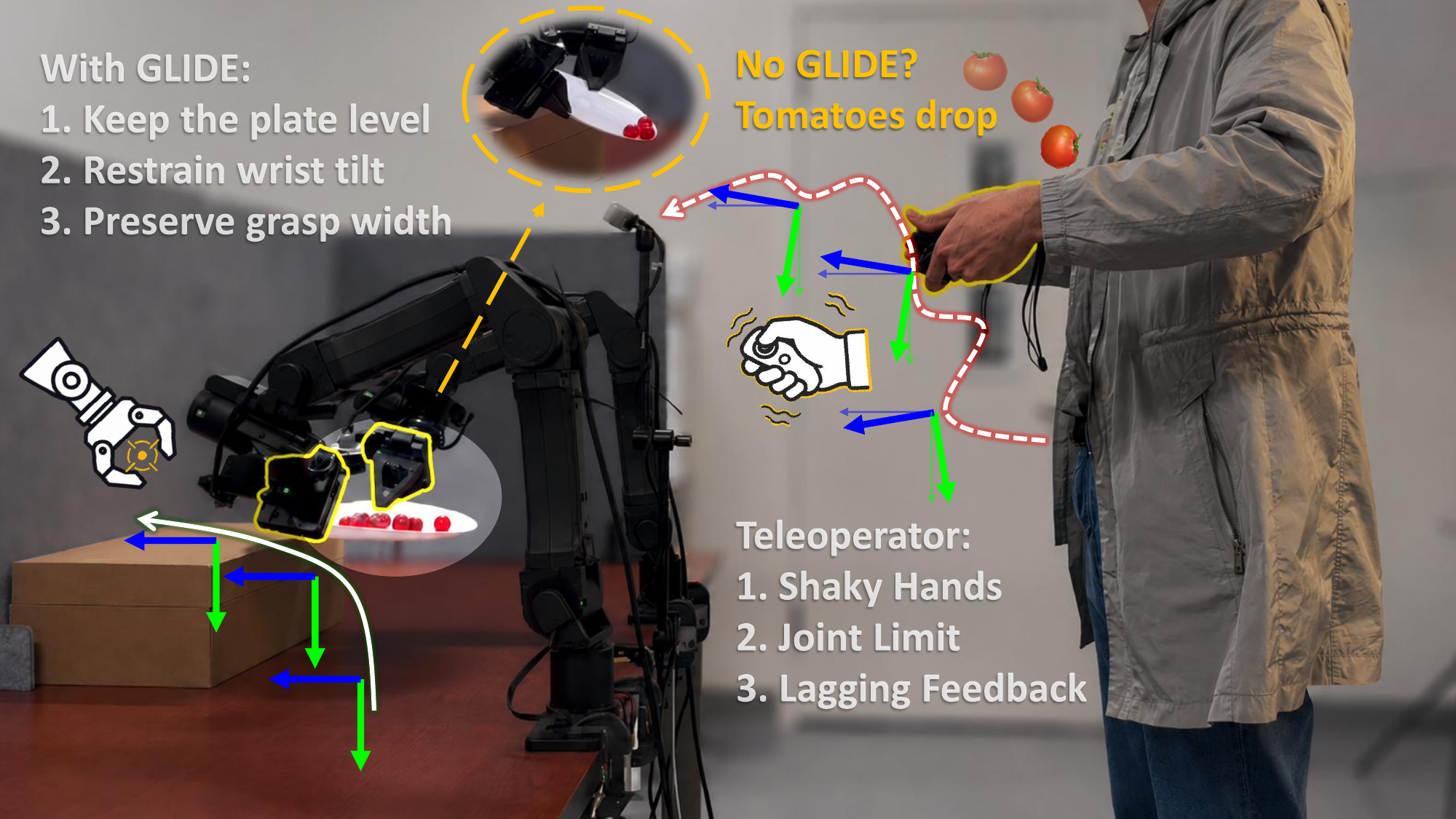}
	\vspace{-1em}
	\caption{\methodname{} converts task descriptions into runtime guardrails that constrain error-prone naive
	teleoperation, enabling reliable data collection and policy learning for
	tasks that are difficult to demonstrate directly.}
	\label{fig:teaser}
\end{wrapfigure}

Behavior cloning from teleoperated demonstrations has become a dominant
paradigm for training robot manipulation
policies~\cite{zhao2023aloha,chi2023diffusionpolicy,fu2025mobilealoha}.
The recipe is straightforward: collect demonstrations, fine-tune a
vision-language-action (VLA) model~\cite{black2024pi0,kim2024openvla}, and deploy.
This approach has produced capable policies across a wide range of
tasks~\cite{zhao2023aloha,fu2025mobilealoha,luo2025hilserl}.
\textbf{Can robots learn tasks that their operators cannot successfully demonstrate?}

This recipe rests on one assumption that is rarely made explicit: a human
operator can successfully demonstrate the task.
For most manipulation benchmarks, this holds: in practice, benchmark tasks tend
to be pick-and-place variations chosen in part because demonstrations are
straightforward to collect at scale.
Modern teleoperation interfaces have further reduced the need for specialized
hardware and enabled remote
operation~\cite{wu2023gello,cheng2024opentelevision,iyer2024openteach}.
Some tasks, however, are simply too demanding to teleoperate reliably, as
\figref{fig:teaser} illustrates.
Consider three such tasks.
A bimanual robot must carry a plate of small round tomatoes between surfaces:
the plate cannot tilt past a critical angle, and any abrupt motion sends
tomatoes rolling.
A robot must pour from a bottle into a wine glass held by a 15-DoF dexterous
hand: success requires stable grasps, bottle-glass alignment, and
controlled wrist motion during the pour.
A robot must handover a marker from one hand to another and place it vertically on a table: the marker must stay
upright, and the gripper must withdraw slowly enough not to knock it over.
Skilled operators fail consistently at all three, leaving behavior cloning with
few or no successful demonstrations.

Even when successes are rare or absent, operators can immediately articulate
what goes wrong: the plate tilts past a critical angle, a tomato rolls to the
rim, the grip shifts too abruptly.
Creating successes may be intractable; inferring what to avoid can still seed
useful guardrails.
\textbf{We propose guardrails and show that inferred task-specific failure
points can make it possible to collect training data, learn a policy, and deploy
it on tasks where naive teleoperation has zero or near-zero success.} The key
effect is not only automation: \methodname{} discovers emergent guardrails that go
beyond domain-expert hardcoded ones, revealing task phases that are
hard to specify a priori.

\textit{Guardrails} encode these constraints as executable code, hard coded or generated by a coding agent, that runs alongside the controller and uses natural
execution events, such as a gripper closing around an object, to infer task
phase.
At each control step, the guardrails enforce the requirements for the inferred
phase of the task and lets execution continue.
\methodname{} builds on this primitive across three stages of the learning pipeline
summarized in \figref{fig:method}. 

During guarded teleoperation, the guardrails constrain the operator's commands
when they risk violating task constraints, such as tilting the plate or drifting
the wine glass during a pour, while leaving the operator in control of task progress.
After refinement, this raises data-collection success from 0--10\% to 70--90\%
across our tasks, making it possible to collect data for tasks that were
previously unreliable to demonstrate.

Through guardrail refinement, each collection session provides feedback for
improving the guardrails themselves.
After each session, \methodname{} revises the guardrails using newly collected
data.
The revisions expose missed phases and unsafe motions that are
difficult to specify before seeing teleoperated trajectories, yielding emergent guardrails
that are absent from the domain-expert hardcoded baseline.

During guarded deployment, the same guardrails wrap the learned policy at test time.
Guarded deployment improves autonomous execution on the tasks where naive policy
outputs are most brittle.
Training on mixed data, including failed attempts partially constrained by the
guardrails, outperforms training on successes only, suggesting these partial
trajectories are informative for learning.

One natural question is how much effort writing guardrails requires.
We show that for a range of tasks, \methodname{} can generate working guardrails from
a task description conditioned on the naive teleoperation framework, then refine it
from data of prior attempts, without manual
tuning or additional
sensing~\cite{ma2024eureka,ma2025gvl,huang2023voxposer,wang2024rlvlmf}.
This makes it practical to deploy \methodname{} on new tasks.

We validate on three tasks: \emph{Tomato plate transfer} and
\emph{Marker handover \& stand} with a bimanual parallel-jaw robot, and
\emph{Wine serving} with a 15-DoF dexterous hand.
Results show consistent improvements in data collection and guarded deployment where naive teleoperation provides few usable demonstrations, and the
learned guardrails go beyond domain-expert hardcoded guardrails. 
This motivates studying the \textit{infeasible demonstration} regime, where
teleoperation success rates are zero or near-zero, and the failure-specification
asymmetry that makes it tractable.

\noindent\textbf{Contributions.}
We make the following contributions:
(1) We propose \methodname, a guardrail framework spanning guarded teleoperation,
guardrail refinement, and guarded deployment.
(2) We show that \methodname{} can discover emergent guardrails from task descriptions, and trajectory feedback, going beyond domain-expert hardcoded guardrails.
(3) We demonstrate 70--90\% guarded data-collection success and 60--70\%
guarded policy success on bimanual and dexterous manipulation tasks where naive
teleoperation provides zero or near-zero successes.


\section{Related Work}
\label{sec:related-work}

\noindent\textbf{Teleoperation interfaces.}
Imitation learning builds on expert demonstrations collected from a
human operator. Leader-follower systems such as ALOHA~\citep{zhao2023aloha}, Mobile
ALOHA~\citep{fu2025mobilealoha}, and GELLO~\citep{wu2023gello} collect precise
demonstrations through paired leader devices; however, they require dedicated
hardware and do not enforce task-level constraints. Vision and XR interfaces
reduce this hardware burden: AnyTeleop~\citep{qin2023anyteleop} relies on camera-based vision tracking, OPEN TEACH~\citep{iyer2024openteach} leverages VR hand tracking, and
Open-TeleVision~\citep{cheng2024opentelevision} enables stereoscopic active visual feedback through a VR headset. Other interfaces add visuotactile hands, force feedback,
haptic cues, or visual-exoskeleton
tracking~\citep{lin2024hato,liu2025factr,ding2025bunnyvisionpro,yang2024ace}.
These interfaces ease data collection, yet they do not check whether a specific command
risks task failure. \methodname{} adds this check at the command interface by
mapping human and policy commands to task-constrained robot actions before
execution.
In concurrent work, Sha et al.~\citep{sha2026copilot} improve teleoperation with a
complementary learned shared-autonomy copilot; unlike their residual assistance policy,
\methodname{} infers executable guardrails from task failures and reuses them for both data collection and policy deployment.

\noindent\textbf{Learning from imperfect demonstrations.}
Prior work typically handles imperfect data during policy
learning.
Human-in-the-loop methods use corrections as supervision or reward:
SIRIUS~\citep{liu2022robot} uses interventions to correct OOD behavior, RLIF~\citep{luo2024rlif} converts interventions into rewards,
and ConRFT~\citep{chen2025conrft} extends reinforcement learning (RL) to VLA policies. Data curation methods
select or weight demonstrations by action or transition diversity, trajectory
influence, or rollout
progress~\citep{belkhale2023dataquality,dass2026datamil,agia2025cupid,ma2025gvl,chen2026topreward}.
DWBC~\citep{xu2022dwbc} reweights behavioral cloning with a discriminator that
separates expert from suboptimal data.
On the algorithmic side,
DAgger~\citep{ross2011reduction} aggregates expert feedback under the learner's
induced state distribution;
Jing et al.~\citep{jing2020soft} treat imperfect demonstrations as soft expert
guidance for RL;
ADVISOR~\citep{weihs2021advisor} adaptively balances imitation and RL losses to
bridge the imitation gap from privileged teachers;
and VRB~\citep{bahl2023affordances} extracts actionable affordances from human
videos to support imitation and RL.
PATO~\citep{dass2023pato} assists operator-guided data collection by executing repetitive
subtasks and querying the human under uncertainty. These methods improve
learning through feedback, selection, or reweighting. \methodname{} intervenes
earlier by constraining commands before execution, then using the resulting
teleoperated trajectories to refine guardrails iteratively.

\noindent\textbf{LLMs for robotics.}
LLMs have accelerated progress in robotics. For planning and
control, SayCan~\citep{ichter2023saycan} grounds language-generated skill
sequences, WildLMa~\citep{qiu2024wildlma} uses LLM-generated plans for
long-horizon loco-manipulation, Code as Policies~\citep{liang2023codeaspolicies}
writes code over perception and robot APIs, and
VoxPoser~\citep{huang2023voxposer} generates 3D value maps for motion planning.
For RL, Eureka~\citep{ma2024eureka}, DrEureka~\citep{ma2024dreureka},
and RL-VLM-F~\citep{wang2024rlvlmf} generate reward signals or domain randomization with LLMs.
Recent coding-agent systems extend this line by synthesizing full manipulation
programs~\citep{fu2026capx}, iteratively debugging and reusing control code from
rollout feedback~\citep{lu2026aspire}, autonomously improving training recipes
and policies in a real-world feedback loop~\citep{xiao2026enpire}, or driving
robots through a browser-based visual interface without robot-specific
fine-tuning~\citep{hu2026via}.
These methods generate complete controllers or improve policy training;
\methodname{} instead writes lightweight guardrails that constrain an existing
operator or learned policy at execution time, targeting tasks where high-quality
demonstrations are hard to collect.


\section{\methodname}
\label{sec:method}

\begin{figure}[t]
	\centering
	\includegraphics[width=0.95\linewidth]{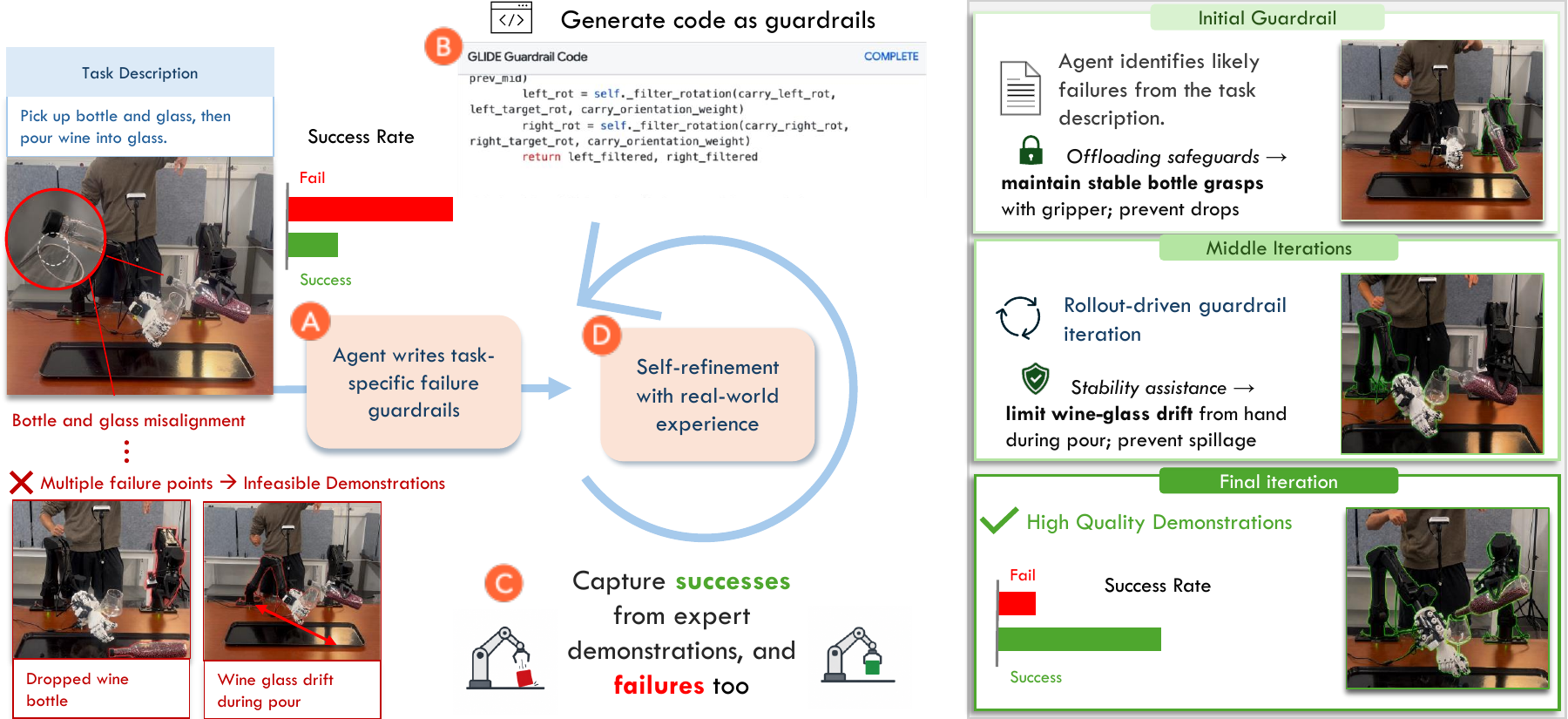}
	\vspace{-0.5em}
	\caption{\textbf{\methodname{} methodology.} (A) From the task description,
            \methodname{} infers task-specific failure points and writes executable guardrails. (B) \methodname{} generates guardrail code.
            (C) During data collection, the guardrails filter proposed commands
            while recording videos, robot states, proposed commands, and executed commands. \methodname{} diagnoses outcomes and failure modes from these recordings.
            (D) \methodname{}
            self-refines the guardrails over iterations; the resulting guardrails
            improve demonstration collection and also guard learned policies at deployment. The animation of this figure can be found at
            \href{https://guardrail-policy.github.io/\#method}{the project webpage}.}
	\label{fig:method}
\end{figure}

We propose \methodname{}, an iterative framework to collect demonstrations
for manipulation tasks that are difficult or impossible with naive teleoperation. \methodname{} generates executable guardrail code
that filters failure-prone commands (\figref{fig:method}). From a standard task
description, \methodname{} infers
potential task-specific failure points and system-state signals, then generates an initial guardrail function over editable
commands such as arm pose, wrist orientation, gripper state, and finger
movement. The user collects demonstrations with active guardrails. After
each round, \methodname{} revises the guardrails using recorded videos, logged state-command tuples, and
episode outcomes. This iterative self-refinement improves the guardrails and
discovers constraints absent from domain-expert hardcoded guardrails. 

We illustrate guardrail refinement from guarded teleoperation through the
\emph{Wine serving} task. In this task, bimanual teleoperation moves the bottle and
glass independently, creating several failure points: glass drift and
uncontrolled bottle tilt during the pour, bottle-glass misalignment, and
end-effector motions that destabilize either grasp. To prevent these failures,
\methodname{} limits hand motion to preserve the glass grasp, amplifies wrist
rotation for controlled pouring, and maintains bottle-glass alignment by
bounding their relative position. This lets the operator focus on timing the
pour instead of correcting grasp constantly.
Appendix~\ref{app:guardrail-design} provides the full guardrail details for all
three tasks.

We describe initialization in \secref{sec:method-guardrails},
trajectory-based self-refinement in \secref{sec:method-refinement}, and deployment for
guarded data collection and policy deployment in
\secref{sec:method-deployment}.

\subsection{\methodname{} Initialization: Seeding Guardrails}
\label{sec:method-guardrails}

At refinement iteration \(k\), \(s_t\) denotes the observed state of the
system. Conditioned on \(s_t\) and the task description \(\ell\), the human
operator proposes action \(a_t\) through the implicit human policy \pihum{}. We
denote the executable guardrails and their teleoperated dataset by \Giter{} and
\Diter{}, respectively. The optimal guardrails \Gopt{} are the guardrails from
the evaluated iteration with the highest success rate for expert demonstration collection. We use \Gde{} for the separately
implemented domain-expert guardrails. The executable guardrails \Giter{} filter
the command:
\begin{equation}
	a_t=\pihum(s_t,\ell),
	\qquad \tilde{a}_t = \Giter(s_t, a_t).
	\label{eq:guardrails-filter}
\end{equation}
where \(\tilde{a}_t\) is sent to the robot. The command \(a_t\) contains
editable targets such as arm pose, wrist orientation, gripper state, and for the dexterous setup, finger
motion.

To initialize \Ginit{}, \methodname{} prompts a coding agent with the task
description \(\ell\) and the conditioning teleoperation code \teleopcode{}, as shown in Fig.~\ref{fig:method} (A):
\begin{equation}
	\Ginit = \operatorname{Write}(\ell,\teleopcode).
	\label{eq:guardrails-init}
\end{equation}
Here, \Ginit{} denotes the initial guardrails before trajectory-based self-refinement.

\methodname{} inspects \teleopcode{} to identify its command interface, editable
actions, and available state signals. The prompt also specifies the task and
hardware, then
asks \methodname{} to identify likely failures, propose constraints, and implement a
filter. For \emph{Wine serving}, it specifies bimanual teleoperation retargeting, a
bottle controlled by the left gripper, a glass cup controlled by the right dexterous
hand, alignment before pouring, spill avoidance, and returning both
objects to the table.
Appendix~\ref{app:implementation} gives the full initialization prompts for all
three tasks.

\noindent\textbf{Task-specific inputs, shared framework.}
The procedure and prompt structure are shared across tasks; task descriptions,
hardware context, and recorded data vary, producing task-specific guardrail code,
phase triggers, and parameters. The refinement prompt is shared across tasks
and rounds, with only its dataset path updated
(Appendix~\ref{app:data-feedback-iterations}).

\subsection{\methodname{} Iterations: Refining Guardrails from Guarded Teleoperation}
\label{sec:method-refinement}

The initial guardrails \Ginit{} can still fail when its constraints are
incomplete, mistimed, or too restrictive, especially for challenging tasks (\tabref{tab:teleop-results}). \methodname{} therefore treats
each collection round as iterative self-refinement: successful trajectories identify useful constraints,
and failed trajectories reveal missing constraints, unsafe motion, or filters that
block setup motion.

At iteration \(k\), the operator collects successful and failed episodes with
\Giter{} active, where the teleoperated action is
\(a_t=\pihum(s_t,\ell)\). For episode \(j\), we record head/wrist video
\(V_j\) and a trajectory \(\tau_j=\{(a_t,s_t,\tilde{a}_t)\}_t\)
of proposed commands, proprioceptive states, and executed commands.
\methodname{} examines both recordings to diagnose the success/failure
outcome \(y_j\) and failure modes, and \Diter{} stores \((V_j,\tau_j,y_j)\).
Humans perform teleoperation, optionally logging outcome labels and intuitive feedback,
with no human-written diagnoses required. The recorded video supports offline diagnosis and the
runtime guardrails use proprioception. For \(k<K\), \methodname{} revises the code:

\begin{equation}
	\Gnext=\operatorname{Revise}(\ell,\Giter,\Diter).
	\label{eq:guardrails-update}
\end{equation}
After evaluating all iterations, we select
\begin{equation}
	k^* \in \arg\max_{0\leq k\leq K}
	\operatorname{SuccessRate}(\Diter),
	\qquad \Gopt = \Gglide{k^*}.
	\label{eq:guardrails-select}
\end{equation}
Thus, \Gopt{} denotes the optimal guardrails.
\algref{alg:glide-refinement} summarizes this loop.

For example, in the \emph{Marker handover \& stand} task, the initial teleoperated trajectories revealed that
the guardrails impeded teleoperation: raw controller motion is much faster than the commanded end-effector speed, and vertical motion repeatedly saturated at the cap. \methodname{} increased filter responsiveness and motion
limits while relaxing excessive orientation and handoff damping. Once this lag
was removed, nine of the next ten episodes brought the marker upright but
toppled it during final gripper opening or retraction. The revision turned
release into an explicit guarded phase that holds position and orientation
while the fingers open and permits only a controlled vertical retreat. In the
following iteration, residual failures showed that this phase still began too
reactively, so \methodname{} delayed and slowed opening and timed the retreat
from when the gripper was actually open. Appendix~\ref{app:data-feedback-iterations}
details this progression; \tabref{tab:marker-guardrail-restrictions} in
Appendix~\ref{app:guardrail-design} formalizes the resulting restrictions.

\begin{figure}[t]
	\centering
	\begin{minipage}{0.96\linewidth}
		\refstepcounter{algorithm}\label{alg:glide-refinement}
		\setcounter{algorithmline}{0}
		\hrule
		\vspace{0.35em}
		\textbf{Algorithm~\thealgorithm} \methodname{} Guardrail Refinement
		\vspace{0.25em}
		\hrule
		\vspace{0.35em}
		\footnotesize
		\begin{tabular}{@{}r@{\hspace{0.5em}}>{\raggedright\arraybackslash}p{0.84\linewidth}@{}}
			\algline{algline:glide-require}{\textbf{Require:} task description $\ell$, naive teleoperation code $\teleopcode$}
			\algline{algline:glide-hyperparameters}{\textbf{Hyperparameters:}
				refinement iterations $K$, episodes per iteration $M$}
			\algline{algline:glide-write}{$\Ginit\gets\operatorname{Write}(\ell,\teleopcode)$
				\algcomment{Generate initial guardrails}}
			\algline{algline:glide-outer-loop}{\textbf{for} $k=0,\ldots,K$ \textbf{do}}
			\algline{algline:glide-init-dataset}{\quad $\Diter \gets \emptyset$}
			\algline{algline:glide-rollout-loop}{\quad \textbf{for} $j=1,\ldots,M$ \textbf{do}}
			\algline{algline:glide-run-vr}{\quad \quad Run teleoperation with \Giter{} active; record video $V_j$}
			\algline{algline:glide-timestep-loop}{\quad \quad \textbf{for} each timestep $t$ \textbf{do}}
			\algline{algline:glide-read-state-action}{\quad \quad \quad Observe system state $s_t$
				and human action $a_t=\pihum(s_t,\ell)$}
			\algline{algline:glide-execute-guardrails}{\quad \quad \quad Execute
				$\tilde{a}_t=\Giter(s_t,a_t)$
				\algcomment{Log states and commands}}
			\algline{algline:glide-timestep-end}{\quad \quad \textbf{end for}}
			\algline{algline:glide-label-outcome}{\quad \quad \methodname{} diagnoses $y_j$ and failure modes from video $V_j$
				and trajectory $\tau_j=\{(a_t,s_t,\tilde{a}_t)\}_t$}
			\algline{algline:glide-update-dataset}{\quad \quad
				$\Diter \gets \Diter \cup
				\{(V_j,\tau_j,y_j)\}$}
			\algline{algline:glide-rollout-end}{\quad \textbf{end for}}
			\algline{algline:glide-revise-comment}{\quad \textbf{if} $k<K$ \textbf{then}}
			\algline{algline:glide-revise}{\quad
				$\Gnext\gets\operatorname{Revise}(\ell,\Giter,\Diter)$
				\algcomment{Revise using successes and failures}}
			\algline{algline:glide-revise-end}{\quad \textbf{end if}}
			\algline{algline:glide-outer-end}{\textbf{end for}}
			\algline{algline:glide-select-final}{Select
				$k^*\in\arg\max_{0\leq k\leq K}\operatorname{SuccessRate}(\Dround{k})$;
				$\Gopt\gets\Gglide{k^*}$}
			\algline{algline:glide-final-data}{Form training pool $\Dpool$ from retained iteration data
				and/or additional data $\Dopt$ collected with \Gopt{} (\secref{sec:method-deployment})}
			\algline{algline:glide-output}{\textbf{Output:} optimal guardrails \Gopt{}
				and episode datasets $\{\Dround{k}\}_{k=0}^{K}$, $\Dpool$}
		\end{tabular}
		\vspace{0.35em}
		\hrule
	\end{minipage}
\end{figure}

\subsection{\methodname{} Deployment: Guarded Data Collection and Policy Deployment}
\label{sec:method-deployment}

Policy training uses a demonstration pool \Dpool{}. For \emph{Tomato plate
transfer} and \emph{Marker handover \& stand}, it accumulates all three evaluated
rounds: $\Dpool=\bigcup_{k=0}^{K}\Dround{k}$, with $K=2$.
For \emph{Wine serving}, the first two rounds yield no successes. Its pool
instead comprises additional data collected with the corrected \Gopt{}:
$\Dpool=\Dopt=\{(V_j,\tau_j,y_j)\}_{j=1}^{M_*}$, excluding earlier refinement episodes.
Outcome labels partition \Dpool{} into successful episodes \Dsucc{} and
imperfect episodes $\Dimpf=\Dpool\setminus\Dsucc$.

We form two distinct policy-training datasets: the success-only set
$\Dtrainsucc=\Dsucc$ and the mixed-quality set
$\Dtrainmix=\Dsucc\cup\Dimpf$. Thus, success-only training uses a subset of
the mixed-quality pool, rather than an equal-sized independent dataset.
The policy is conditioned on the system state
$s_t$ and language task description $\ell$, with executed command
$\tilde{a}_t$ as the action target. We train the policies with flow-matching imitation-learning objective~\citep{black2024pi0}:
\begin{equation}
	\hat{\pi}_{q}
	= \arg\max_{\pi}
	\mathbb{E}_{(s_t,\ell,\tilde{a}_t)\sim\mathcal{D}_{\mathrm{train}}^{q}}
	\left[\log \pi(\tilde{a}_t\mid s_t,\ell)\right],
	\qquad q\in\{\mathrm{succ},\mathrm{mix}\}.
	\label{eq:guarded-bc}
\end{equation}
At deployment, the same guardrails \Gopt{} filter the policy's
predicted action at execution time:
\begin{equation}
	\tilde{a}_t=\Gopt\!\left(s_t,\hat{\pi}_{q}(s_t,\ell)\right).
	\label{eq:guarded-deployment}
\end{equation}
Thus \Gopt{} supports both demonstration collection and policy execution.


\section{Experiments}
\label{sec:experiments}

\subsection{Evaluation and Baselines}
\label{sec:exp-setup}

\newenvironment{resultlist}{%
	\begin{list}{$\bullet$}{%
		\setlength{\leftmargin}{1.25em}%
		\setlength{\labelwidth}{0.75em}%
		\setlength{\labelsep}{0.35em}%
		\setlength{\itemsep}{0.2em}%
		\setlength{\parsep}{0pt}%
		\setlength{\topsep}{0.25em}%
		\setlength{\partopsep}{0pt}%
	}%
}{%
	\end{list}%
}

\noindent Robot setup and teleoperation details appear in
Appendix~\ref{app:robot-setup} and \figref{fig:teleop-setup}.

\noindent\textbf{Gripper Tasks.}
We evaluate two gripper tasks that require stable and coordinated object
transport. 

\begin{resultlist}
	\item \emph{Tomato plate transfer}: The robot must use both
	grippers to grasp and move a plate with eight round tomatoes from the tabletop to an
	elevated surface. Successful execution requires keeping the plate stable
	through lift and placement so tomatoes are not spilled.

	\item \emph{Marker handover \& stand}: The robot must grasp a marker,
	transfer it to the opposing gripper, place it upright on the tabletop, and
	withdraw without knocking it over. Successful execution requires coordinated
	handoff and gentle release.
\end{resultlist}

\noindent\textbf{Dexterous Task.}
We also evaluate one task that involves a dexterous hand.

\begin{resultlist}
	\item \emph{Wine serving}: The robot must use the left gripper to
	grasp the wine bottle and the right dexterous hand to grasp the glass cup, then
	coordinate the two end effectors to pour wine from the bottle into the glass.
	Successful execution requires stable grasps, controlled bottle tilt, and a
	stable glass pose during pouring.
\end{resultlist}

\noindent\textbf{Baselines.}
We use separate baselines for data collection and policy execution.

\begin{resultlist}
	\item \texttt{OpenTV}~\citep{cheng2024opentelevision}
	(\emph{unguarded teleoperation}): maps VR controller readings to robot joint
	commands through inverse kinematics and executes them without task-specific
	guardrail filtering.

	\item \texttt{Manual} (\emph{domain-expert hardcoded guardrails}, \Gde{}):
	the data-collection guardrail baseline. It filters end-effector commands with
	domain-expert hardcoded task constraints before execution.

	\item \textbf{$\pi_{\mathrm{0.5}}$}~\citep{black2024pi0}
	(\emph{unguarded policy execution}): executes predicted actions directly. This gives a controlled comparison with \Gopt{} under both
	success-only and mixed-quality \methodname{} training data. We also evaluate
	policies trained on unguarded OpenTV demonstrations (\emph{Vanilla}) with
	and without \Gopt{} at deployment.
\end{resultlist}

\noindent\textbf{Metrics.}
We report three metrics in \tabref{tab:teleop-results} and
\tabref{tab:deployment-results}.

\begin{resultlist}
	\item \emph{Success rate}: fraction out of ten trials that satisfy
	the task-specific completion criterion.

	\item \emph{Average tomatoes left}: for \emph{Tomato
	plate transfer} task only, which evaluate transport stability, measured as the average number of tomatoes remaining in the plate from the initial eight.

	\item \emph{Throughput per minute}: execution efficiency for \emph{Marker
	handover \& stand} and \emph{Wine serving}, measured as number of successful completions
	per elapsed minute.
\end{resultlist}

\subsection{\methodname{}-Enabled Data Collection}
\label{sec:results-data-collection}

During data collection, \emph{OpenTV}~\citep{cheng2024opentelevision} executes
raw VR-retargeted commands, \emph{Manual} applies \Gde{}, and \methodname{} generates
guardrails without using domain-expert task restrictions. At each \methodname{}
iteration, we collect ten episodes and revise the guardrails from the resulting
guarded teleoperation. Each baseline and each reported \methodname{} round
is evaluated over $M=10$ trials.
The optimal \Gopt{} filters are phase-triggered command restrictions:
coupled carry for plates, handoff/release stabilization for markers, and bottle-glass alignment constraints for wine serving.
Appendix~\ref{app:guardrail-design} gives the full details of the guardrails that are \methodname{}-only, shared, and domain-expert-only restrictions.

\begin{table}[t]
	\centering
	\caption{\textbf{Zero-to-one on collecting expert demonstrations.} Raw
	OpenTV yields zero or one success out of ten for
	\tomatoemoji{}~\texttt{Tomato plate transfer},
	\markeremoji{}~\texttt{Marker handover \& stand}, and
	\wineemoji{}~\texttt{Wine serving}, while \methodname{} with the optimal guardrails \Gopt{}
	reaches 70--90\% success. Succ.: success rate; Avg. \tomatoemoji{} left:
	average tomatoes left over all episodes
	(from 8); Thr./min: throughput per minute.
	$\uparrow$ indicates higher the better ; bold marks the best value in each column.}
	\label{tab:teleop-results}
	\setlength{\tabcolsep}{1.8pt}
	\begin{tabular}{l
		>{\columncolor{tasktomatobg}}c >{\columncolor{tasktomatobg}}c
		>{\columncolor{taskmarkerbg}}c >{\columncolor{taskmarkerbg}}c
		>{\columncolor{taskwinebg}}c >{\columncolor{taskwinebg}}c}
		\toprule
		\textbf{Method} &
		\multicolumn{2}{>{\columncolor{tasktomatobg}}c}{\textbf{\tomatoemoji{}~Tomato plate transfer}} &
		\multicolumn{2}{>{\columncolor{taskmarkerbg}}c}{\textbf{\markeremoji{}~Marker handover \& stand}} &
		\multicolumn{2}{>{\columncolor{taskwinebg}}c}{\textbf{\wineemoji{}~Wine serving}} \\
		\cmidrule(lr){2-3}\cmidrule(lr){4-5}\cmidrule(l){6-7}
		& \,Succ.~$\uparrow$ & Avg. \tomatoemoji{} left~$\uparrow$ &
		\,Succ.~$\uparrow$ & Thr./min~$\uparrow$ &
		\,Succ.~$\uparrow$ & Thr./min~$\uparrow$ \\
		\midrule
		OpenTV~\citep{cheng2024opentelevision} & 0\% & 0.3 & 10\% & 0.5 & 0\% & 0.0 \\
		Manual & 20\% & 5.4 & 60\% & 2.8 & 60\% & 0.6 \\
		\midrule
		\Ginit{} & 60\% & \textbf{7.4} & 70\% & 2.1 & 0\% & 0.0 \\
		\Gglide{1} & 60\% & 7.1 & 60\% & 2.6 & 0\% & 0.0 \\
		\Gopt{} & \textbf{70\%} & 6.7 & \textbf{90\%} &
		\textbf{3.9} & \textbf{90\%} & \textbf{0.9} \\
		\bottomrule
	\end{tabular}%
\end{table}

\noindent\textbf{\methodname{} improves expert demonstration success
over naive and domain-expert hardcoded guardrails (\tabref{tab:teleop-results}).}
OpenTV reaches only 0\%, 10\%, and 0\% success on \emph{Tomato plate transfer},
\emph{Marker handover \& stand}, and \emph{Wine serving} tasks, while Manual reaches
20\%, 60\%, and 60\%. With \Gopt{}, \methodname{} raises success to 70\%, 90\%, and
90\%. The successful guardrails address the dominant failures with coupled plate
carry after grasp, close-range marker handoff and release support, and
phase-aware bottle tilt with bottle-mouth alignment and hand grasp limits.

\noindent\textbf{\methodname{} improves task-specific demonstration quality and
efficiency (\tabref{tab:teleop-results}).}
The non-binary metrics show the same pattern. In \emph{Tomato plate transfer},
\methodname{} keeps more than 6 of 8 tomatoes on the plate by coupling the two
grippers and bounding height, width, and transport motion during carry. In
\emph{Marker handover \& stand}, smooth approach, handoff damping, and release
stabilization improve throughput from 0.5 to 3.9 successful completions per minute. In
\emph{Wine serving}, phase-aware tilt and bottle-mouth alignment
assistance help the operator complete pours, boosting throughput from 0 to 0.9 successful completions
per minute.

\noindent\textbf{Trajectory feedback improves \methodname{} performance
(\tabref{tab:teleop-results}).}
Self-refinement does not monotonically improve \methodname, but final reliability goes up across tasks.
\emph{Tomato plate transfer} rises from 60\% to 70\%
success, \emph{Marker handover \& stand} reaches 90\% after an intermediate
drop, and \emph{Wine serving} improves from 0\% in the first two rounds to 90\% after
the second refinement. Guarded teleoperation in
Appendix~\ref{app:data-feedback-iterations} show what changed: plate transport
became a coupled shared-midpoint carry, marker standing gained an explicit gentle release
phase, and wine serving learned to guide the bottle-glass alignment and amplify bottle rotation during pouring.

\begin{figure}[t]
	\centering
	\includegraphics[width=0.95\linewidth]{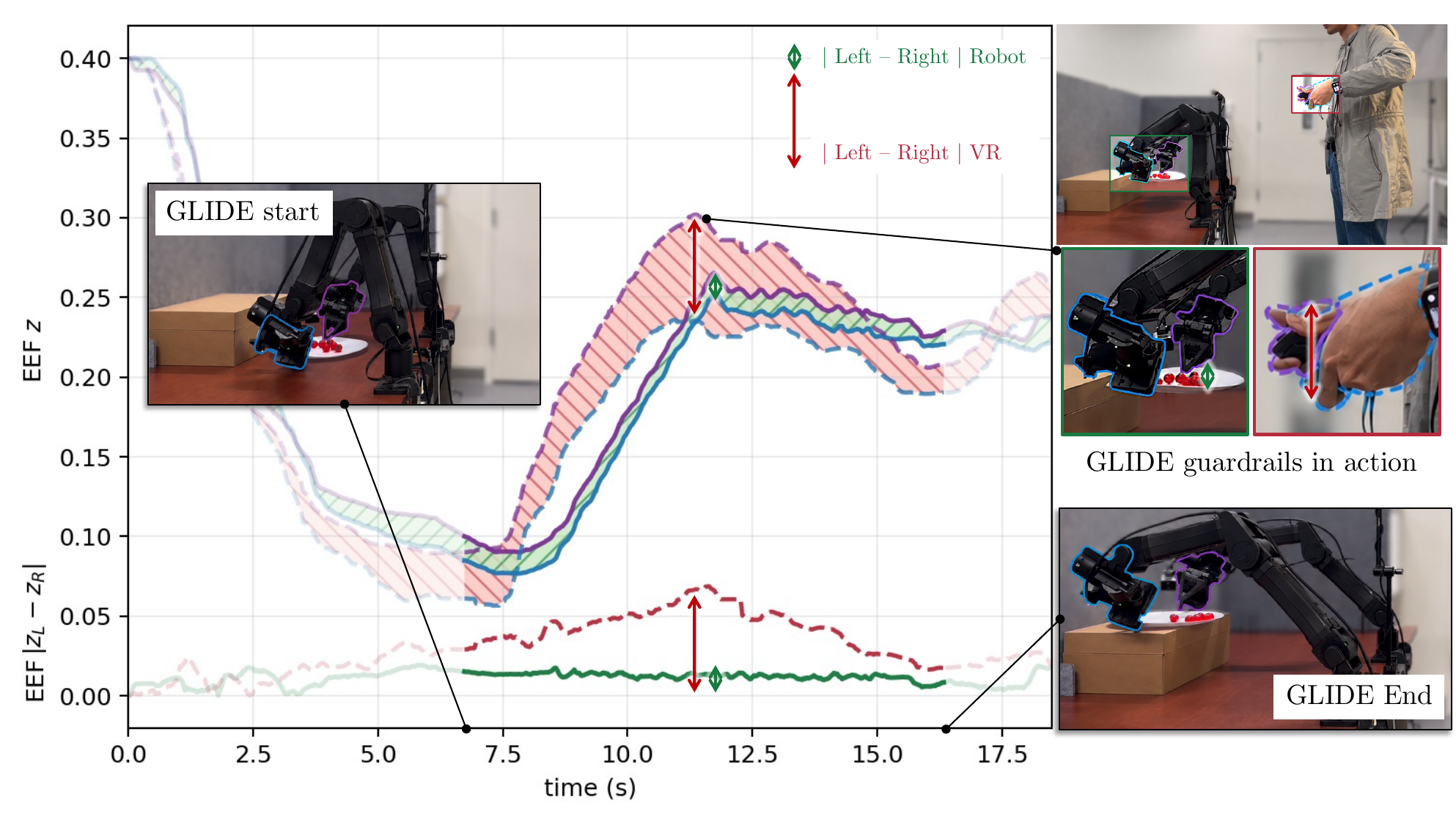}
	\vspace{-1em}
	\caption{\textbf{\methodname{} on \emph{Tomato plate
	transfer}.} In naive teleoperation, discrepancies between the two VR
	controllers can create an inter-arm height mismatch that tilts the plate
	during transport. \methodname{} activates during the carry phase and constrains the
	end-effector height difference, preserving plate stability as the robot lifts
	and transfers the plate. More results can be found at
            \href{https://guardrail-policy.github.io/\#case-study-tomato}{the project webpage}.}
	\label{fig:qualitative}
\end{figure}

\noindent\textbf{The advantage of \methodname{} comes from emergent, phase-specific
\Gopt{} guardrails that go beyond domain-expert hardcoded guardrails
(\figref{fig:qualitative}; Appendix~\ref{app:guardrail-design}).}
\methodname{} is not just reproducing the domain-expert hardcoded baseline. It discovers emergent constraints: adding
coupled plate-carry motion and paired gripper control, marker handoff and
release stabilization, and wine bottle-over-glass alignment with grasp and tilt
guards. The baselines overlap on narrower restrictions,
such as plate leveling, grasp-width preservation, and rotation suppression. Appendix~\ref{app:guardrail-design} separates these shared restrictions from emergent ones discovered
through trajectory feedback.

\subsection{\methodname{}-Assisted Policy Deployment}
\label{sec:results-policy-execution}

For \emph{Tomato plate transfer} and \emph{Marker handover \& stand}, training
reuses all 30 episodes from the three \methodname{} rounds, including 19 and 22
successes, respectively. For \emph{Wine serving}, we collect 60 separate episodes
with the corrected \Gopt{}: 30 successes and 30 imperfect demonstrations.
We fine-tune $\pi_{\mathrm{0.5}}$ on the success-only subset \Dtrainsucc{} or
complete mixed-quality pool \Dtrainmix{}; guarded/unguarded pairs use the same
trained policy. \emph{Vanilla} policies separately use unguarded OpenTV data. Neither OpenTV nor
Manual data enter the \methodname{} pools. Unguarded policies execute raw
predictions; guarded policies apply the same \Gopt{} used for teleoperation
(Appendix~\ref{app:guardrail-design}).

\begin{table}[t]
	\centering
	\caption{\textbf{Training data and runtime guardrails both contribute to policy performance.}
	$\pi_{\mathrm{0.5}}$~\citep{black2024pi0} policies use success-only
	\successdata{} or mixed success/imperfect \mixeddata{} \methodname{}
	demonstrations, or Vanilla (OpenTV) demonstrations \vanilladata{}, and are deployed
	with or without \Gopt{}. Each row reports ten evaluation trials.
	Succ.: success rate; Avg. \tomatoemoji{} left: tomatoes remaining from eight,
	averaged over all trials; Thr./min: successful completions per minute.
	Higher is better; bold marks column maxima.}
	\label{tab:deployment-results}
	\setlength{\tabcolsep}{1.4pt}
	\begin{tabular}{lc
		>{\columncolor{tasktomatobg}}c >{\columncolor{tasktomatobg}}c
		>{\columncolor{taskmarkerbg}}c >{\columncolor{taskmarkerbg}}c
		>{\columncolor{taskwinebg}}c >{\columncolor{taskwinebg}}c}
		\toprule
				\multicolumn{2}{l}{\textbf{\methodname{}-$\pi_{\mathrm{0.5}}$ Policy}} &

		\multicolumn{2}{>{\columncolor{tasktomatobg}}c}{\textbf{\tomatoemoji{}~Tomato plate transfer}} &
		\multicolumn{2}{>{\columncolor{taskmarkerbg}}c}{\textbf{\markeremoji{}~Marker handover \& stand}} &
		\multicolumn{2}{>{\columncolor{taskwinebg}}c}{\textbf{\wineemoji{}~Wine serving}} \\
		\cmidrule(lr){1-2}\cmidrule(lr){3-4}\cmidrule(lr){5-6}\cmidrule(l){7-8}
		Data & Guarded? & Succ.~$\uparrow$ & Avg. \tomatoemoji{} left~$\uparrow$ &
		Succ.~$\uparrow$ & Thr./min~$\uparrow$ &
		Succ.~$\uparrow$ & Thr./min~$\uparrow$ \\
		\midrule
		\vanilladata & \guardno & 0\% & 0.1 & 0\% & 0.0 & 0\% & 0.0 \\
		\vanilladata & \guardyes & 20\% & 2.4 & 20\% & 1.1 & 0\% & 0.0 \\
		\midrule
		\successdata & \guardno & 0\% & 3.0 & 20\% & 1.0 & 50\% & \textbf{0.6} \\
		\successdata & \guardyes & 60\% & 6.2 & 50\% & 2.3 & 40\% & 0.4 \\
		\mixeddata & \guardno & 0\% & 3.5 & 40\% & 2.0 & 40\% & 0.5 \\
		\mixeddata & \guardyes & \textbf{70\%} & \textbf{6.3} &
		\textbf{60\%} & \textbf{2.9} & \textbf{60\%} & 0.5 \\
		\bottomrule
	\end{tabular}%
    \vspace{-1em}
\end{table}

\noindent\textbf{\methodname{} improves autonomous execution where the raw policy is
most brittle (\tabref{tab:deployment-results}).}
With success-only training data, adding the guardrails increases
\emph{Tomato plate transfer} from 0\% to 60\% success and
\emph{Marker handover \& stand} from 20\% to 50\% success, showing that
execution-time filtering can recover policies whose raw action predictions are brittle.

\noindent\textbf{Guarded deployment allows policies to benefit from
mixed-quality demonstrations (\tabref{tab:deployment-results}).}
The mixed-data guarded policy has the best success profile: 70\% on
\emph{Tomato plate transfer}, 60\% on \emph{Marker handover \& stand}, and 60\%
on \emph{Wine serving}, with the highest tomato retention and marker throughput.
With the same mixed-data policy and task order, removing \Gopt{} yields
0\%, 40\%, and 40\% success. This comparison isolates the contribution of
runtime filtering for a fixed policy trained on mixed \methodname{} data.
For \emph{Wine serving}, unguarded success-only execution
reaches 50\% success versus 40\% with \Gopt{}, but mixed-quality
training with guarded execution reaches 60\%. These results support the design of \secref{sec:method-deployment}: imperfect demonstrations can add useful
training data when \Gopt{} remains active.

\noindent\textbf{Runtime guardrails alone are insufficient (\tabref{tab:deployment-results}).}
Vanilla-data policies achieve 0\% success on all tasks without guardrails and
20\%, 20\%, and 0\% with \Gopt{}, below the mixed-data guarded policies.
These comparisons indicate the importance of both assisted collection and
runtime filtering.


\section{Conclusion and Limitations}
\label{sec:conclusion}

This paper studies manipulation tasks where naive VR teleoperation provides few
reliable demonstrations. \methodname{} places executable
guardrails at the command interface, refines them from trajectory feedback, and
uses the optimal guardrails during data collection and policy deployment.
Across \emph{Tomato plate transfer}, \emph{Marker handover \& stand}, and
\emph{Wine serving}, \methodname{} improves data collection over naive
teleoperation and domain-expert hardcoded guardrails. The results show that
\methodname{} discovers guardrails beyond domain-expert hardcoding and lets
policies benefit from mixed-quality demonstrations. These results suggest a
promising direction for useful scaling of robotics data.

\paragraph{Limitations.}
Our results are encouraging but leave several opportunities for future work.
First, the runtime guardrails use only proprioception to filter commands, although
the coding agent uses recorded video and numerical trajectories for offline
diagnosis. Adding visual and tactile signals at runtime could cover slipping
grasps, object motion, and liquid state. Second, guardrails
already condition on robot state, but uncertainty or task-phase estimates could
improve intervention timing. Future studies can extend \methodname{} to long-horizon tasks, mobile manipulation, and with more complex robot systems.


\clearpage
\acknowledgments{We thank Leo Lin, Shivansh Patel and Jay Moon for their help on setting up the CRAFT hand, we thank Basavasagar Patil for helpful feedback on the draft.}

\bibliography{example}  

@Misc{black2024pi0,
  title         = {$\pi_0$: A Vision-Language-Action Flow Model for General Robot Control},
  author        = {Kevin Black and Noah Brown and Danny Driess and Adnan Esmail and Michael Equi and Chelsea Finn and Niccolo Fusai and Lachy Groom and Karol Hausman and Brian Ichter and Szymon Jakubczak and Tim Jones and Liyiming Ke and Sergey Levine and Adrian Li-Bell and Mohith Mothukuri and Suraj Nair and Karl Pertsch and Lucy Xiaoyang Shi and James Tanner and Quan Vuong and Anna Walling and Haohuan Wang and Ury Zhilinsky},
  year          = {2024},
  eprint        = {2410.24164},
  archivePrefix = {arXiv},
  primaryClass  = {cs.LG},
  doi           = {10.48550/arXiv.2410.24164},
  url           = {https://arxiv.org/abs/2410.24164}
}

@InProceedings{kim2024openvla,
  title     = {{OpenVLA}: An Open-Source Vision-Language-Action Model},
  author    = {{Moo Jin} Kim and Karl Pertsch and Siddharth Karamcheti and Ted Xiao and Ashwin Balakrishna and Suraj Nair and Rafael Rafailov and Ethan P Foster and Pannag R Sanketi and Quan Vuong and Thomas Kollar and Benjamin Burchfiel and Russ Tedrake and Dorsa Sadigh and Sergey Levine and Percy Liang and Chelsea Finn},
  booktitle = {Proceedings of The 8th Conference on Robot Learning},
  pages     = {2679--2713},
  year      = {2025},
  volume    = {270},
  series    = {Proceedings of Machine Learning Research},
  publisher = {PMLR},
  url       = {https://proceedings.mlr.press/v270/kim25c.html}
}

@InProceedings{chi2023diffusionpolicy,
  title     = {Diffusion Policy: Visuomotor Policy Learning via Action Diffusion},
  author    = {Cheng Chi and Siyuan Feng and Yilun Du and Zhenjia Xu and Eric Cousineau and Benjamin CM Burchfiel and Shuran Song},
  booktitle = {Proceedings of Robotics: Science and Systems},
  year      = {2023},
  address   = {Daegu, Republic of Korea},
  month     = {July},
  doi       = {10.15607/RSS.2023.XIX.026},
  url       = {https://roboticsproceedings.org/rss19/p026.html}
}

@Misc{zhao2023aloha,
  title         = {Learning Fine-Grained Bimanual Manipulation with Low-Cost Hardware},
  author        = {Tony Z. Zhao and Vikash Kumar and Sergey Levine and Chelsea Finn},
  year          = {2023},
  eprint        = {2304.13705},
  archivePrefix = {arXiv},
  primaryClass  = {cs.RO},
  doi           = {10.48550/arXiv.2304.13705},
  url           = {https://arxiv.org/abs/2304.13705}
}

@InProceedings{fu2025mobilealoha,
  title     = {Mobile {ALOHA}: Learning Bimanual Mobile Manipulation using Low-Cost Whole-Body Teleoperation},
  author    = {Zipeng Fu and Tony Z. Zhao and Chelsea Finn},
  booktitle = {Proceedings of The 8th Conference on Robot Learning},
  pages     = {4066--4083},
  year      = {2025},
  volume    = {270},
  series    = {Proceedings of Machine Learning Research},
  publisher = {PMLR},
  url       = {https://proceedings.mlr.press/v270/fu25b.html}
}

@Misc{wu2023gello,
  title         = {{GELLO}: A General, Low-Cost, and Intuitive Teleoperation Framework for Robot Manipulators},
  author        = {Philipp Wu and Yide Shentu and Zhongke Yi and Xingyu Lin and Pieter Abbeel},
  year          = {2023},
  eprint        = {2309.13037},
  archivePrefix = {arXiv},
  primaryClass  = {cs.RO},
  doi           = {10.48550/arXiv.2309.13037},
  url           = {https://arxiv.org/abs/2309.13037}
}

@InProceedings{lin2024hato,
  title         = {Learning Visuotactile Skills With Two Multifingered Hands},
  author        = {Toru Lin and Yu Zhang and Qiyang Li and Haozhi Qi and Brent Yi and Sergey Levine and Jitendra Malik},
  booktitle     = {2025 IEEE International Conference on Robotics and Automation (ICRA)},
  pages         = {5637--5643},
  year          = {2025},
  doi           = {10.1109/ICRA55743.2025.11128180},
  eprint        = {2404.16823},
  archivePrefix = {arXiv},
  primaryClass  = {cs.RO},
  url           = {https://arxiv.org/abs/2404.16823}
}

@InProceedings{liu2025factr,
  title     = {{FACTR}: Force-Attending Curriculum Training for Contact-Rich Policy Learning},
  author    = {Jason Jingzhou Liu and Yulong Li and Kenneth Shaw and Tony Tao and Ruslan Salakhutdinov and Deepak Pathak},
  booktitle = {Proceedings of Robotics: Science and Systems},
  year      = {2025},
  doi       = {10.15607/RSS.2025.XXI.079},
  url       = {https://www.roboticsproceedings.org/rss21/p079.html}
}

@InProceedings{yang2024ace,
  title     = {{ACE}: A Cross-platform and visual-Exoskeletons System for Low-Cost Dexterous Teleoperation},
  author    = {Shiqi Yang and Minghuan Liu and Yuzhe Qin and Runyu Ding and Jialong Li and Xuxin Cheng and Ruihan Yang and Sha Yi and Xiaolong Wang},
  booktitle = {Proceedings of The 8th Conference on Robot Learning},
  pages     = {4895--4911},
  year      = {2025},
  volume    = {270},
  series    = {Proceedings of Machine Learning Research},
  publisher = {PMLR},
  url       = {https://proceedings.mlr.press/v270/yang25d.html}
}

@InProceedings{qin2023anyteleop,
  title     = {AnyTeleop: A General Vision-Based Dexterous Robot Arm-Hand Teleoperation System},
  author    = {Yuzhe Qin and Wei Yang and Binghao Huang and Karl Van Wyk and Hao Su and Xiaolong Wang and Yu-Wei Chao and Dieter Fox},
  booktitle = {Robotics: Science and Systems},
  year      = {2023},
  url       = {https://roboticsconference.org/2023/program/papers/015}
}

@Misc{iyer2024openteach,
  title         = {{OPEN TEACH}: A Versatile Teleoperation System for Robotic Manipulation},
  author        = {Aadhithya Iyer and Zhuoran Peng and Yinlong Dai and Irmak Guzey and Siddhant Haldar and Soumith Chintala and Lerrel Pinto},
  year          = {2024},
  eprint        = {2403.07870},
  archivePrefix = {arXiv},
  primaryClass  = {cs.RO},
  doi           = {10.48550/arXiv.2403.07870},
  url           = {https://arxiv.org/abs/2403.07870}
}

@InProceedings{cheng2024opentelevision,
  title     = {Open-TeleVision: Teleoperation with Immersive Active Visual Feedback},
  author    = {Xuxin Cheng and Jialong Li and Shiqi Yang and Ge Yang and Xiaolong Wang},
  booktitle = {Proceedings of The 8th Conference on Robot Learning},
  pages     = {2729--2749},
  year      = {2025},
  volume    = {270},
  series    = {Proceedings of Machine Learning Research},
  publisher = {PMLR},
  url       = {https://proceedings.mlr.press/v270/cheng25b.html}
}

@InProceedings{ding2025bunnyvisionpro,
  title     = {Bunny-VisionPro: Real-Time Bimanual Dexterous Teleoperation for Imitation Learning},
  author    = {Runyu Ding and Yuzhe Qin and Jiyue Zhu and Chengzhe Jia and Shiqi Yang and Ruihan Yang and Xiaojuan Qi and Xiaolong Wang},
  booktitle = {2025 IEEE/RSJ International Conference on Intelligent Robots and Systems (IROS)},
  pages     = {12248--12255},
  year      = {2025},
  doi       = {10.1109/IROS60139.2025.11247017},
  url       = {https://ieeexplore.ieee.org/document/11247017}
}

@Misc{lin2026craft,
  title         = {{CRAFT}: A Tendon-Driven Hand with Hybrid Hard-Soft Compliance},
  author        = {Leo Lin and Shivansh Patel and Jay Moon and Svetlana Lazebnik and Unnat Jain},
  year          = {2026},
  eprint        = {2603.12120},
  archivePrefix = {arXiv},
  primaryClass  = {cs.RO},
  doi           = {10.48550/arXiv.2603.12120},
  url           = {https://arxiv.org/abs/2603.12120}
}

@Article{luo2025hilserl,
  title   = {Precise and Dexterous Robotic Manipulation via Human-in-the-Loop Reinforcement Learning},
  author  = {Jianlan Luo and Charles Xu and Jeffrey Wu and Sergey Levine},
  journal = {Science Robotics},
  year    = {2025},
  volume  = {10},
  number  = {105},
  pages   = {eads5033},
  doi     = {10.1126/scirobotics.ads5033},
  url     = {https://www.science.org/doi/10.1126/scirobotics.ads5033}
}

@InProceedings{luo2024rlif,
  title     = {{RLIF}: Interactive Imitation Learning as Reinforcement Learning},
  author    = {Jianlan Luo and Perry Dong and Yuexiang Zhai and Yi Ma and Sergey Levine},
  booktitle = {International Conference on Learning Representations},
  year      = {2024},
  url       = {https://arxiv.org/abs/2311.12996}
}

@InProceedings{chen2025conrft,
  title     = {{ConRFT}: A Reinforced Fine-tuning Method for {VLA} Models via Consistency Policy},
  author    = {Yuhui Chen and Shuai Tian and Shugao Liu and Yingting Zhou and Haoran Li and Dongbin Zhao},
  booktitle = {Robotics: Science and Systems},
  year      = {2025},
  url       = {https://roboticsconference.org/program/papers/19}
}

@InProceedings{belkhale2023dataquality,
  title     = {Data Quality in Imitation Learning},
  author    = {Suneel Belkhale and Yuchen Cui and Dorsa Sadigh},
  booktitle = {Advances in Neural Information Processing Systems},
  volume    = {36},
  year      = {2023},
  url       = {https://proceedings.neurips.cc/paper_files/paper/2023/hash/fe692980c5d9732cf153ce27947653a7-Abstract-Conference.html}
}

@InProceedings{dass2026datamil,
  title     = {{DataMIL}: Selecting Data for Robot Imitation Learning with Datamodels},
  author    = {Shivin Dass and Alaa Khaddaj and Logan Engstrom and Aleksander Madry and Andrew Ilyas and Roberto Mart{\'i}n-Mart{\'i}n},
  booktitle = {International Conference on Learning Representations},
  year      = {2026},
  url       = {https://openreview.net/forum?id=AcTsKglDdh}
}

@InProceedings{agia2025cupid,
  title     = {{CUPID}: Curating Data your Robot Loves with Influence Functions},
  author    = {Christopher Agia and Rohan Sinha and Jingyun Yang and Rika Antonova and Marco Pavone and Haruki Nishimura and Masha Itkina and Jeannette Bohg},
  booktitle = {Proceedings of The 9th Conference on Robot Learning},
  pages     = {2907--2932},
  year      = {2025},
  volume    = {305},
  series    = {Proceedings of Machine Learning Research},
  publisher = {PMLR},
  url       = {https://proceedings.mlr.press/v305/agia25a.html}
}

@InProceedings{ma2025gvl,
  title     = {Vision Language Models are In-Context Value Learners},
  author    = {Yecheng Jason Ma and Joey Hejna and Ayzaan Wahid and Chuyuan Fu and Dhruv Shah and Jacky Liang and Zhuo Xu and Sean Kirmani and Peng Xu and Danny Driess and Ted Xiao and Jonathan Tompson and Osbert Bastani and Dinesh Jayaraman and Wenhao Yu and Tingnan Zhang and Dorsa Sadigh and Fei Xia},
  booktitle = {International Conference on Learning Representations},
  year      = {2025},
  url       = {https://openreview.net/forum?id=friHAl5ofG}
}

@Misc{chen2026topreward,
  title         = {{TOPReward}: Token Probabilities as Hidden Zero-Shot Rewards for Robotics},
  author        = {Shirui Chen and Cole Harrison and Ying-Chun Lee and Angela Jin Yang and Zhongzheng Ren and Lillian J. Ratliff and Jiafei Duan and Dieter Fox and Ranjay Krishna},
  year          = {2026},
  eprint        = {2602.19313},
  archivePrefix = {arXiv},
  primaryClass  = {cs.RO},
  doi           = {10.48550/arXiv.2602.19313},
  url           = {https://arxiv.org/abs/2602.19313}
}

@InProceedings{xu2022dwbc,
  title     = {Discriminator-Weighted Offline Imitation Learning from Suboptimal Demonstrations},
  author    = {Haoran Xu and Xianyuan Zhan and Honglei Yin and Huiling Qin},
  booktitle = {Proceedings of the 39th International Conference on Machine Learning},
  pages     = {24725--24742},
  year      = {2022},
  volume    = {162},
  series    = {Proceedings of Machine Learning Research},
  publisher = {PMLR},
  url       = {https://proceedings.mlr.press/v162/xu22l.html}
}

@InProceedings{ross2011reduction,
  title     = {A Reduction of Imitation Learning and Structured Prediction to No-Regret Online Learning},
  author    = {St{\'e}phane Ross and Geoffrey Gordon and Drew Bagnell},
  booktitle = {Proceedings of the Fourteenth International Conference on Artificial Intelligence and Statistics},
  pages     = {627--635},
  year      = {2011},
  volume    = {15},
  series    = {Proceedings of Machine Learning Research},
  publisher = {PMLR},
  url       = {https://proceedings.mlr.press/v15/ross11a.html}
}

@InProceedings{jing2020soft,
  title     = {Reinforcement Learning from Imperfect Demonstrations under Soft Expert Guidance},
  author    = {Mingxuan Jing and Xiaojian Ma and Wenbing Huang and Fuchun Sun and Chao Yang and Bin Fang and Huaping Liu},
  booktitle = {Proceedings of the AAAI Conference on Artificial Intelligence},
  volume    = {34},
  number    = {04},
  pages     = {5109--5116},
  year      = {2020},
  doi       = {10.1609/aaai.v34i04.5953},
  url       = {https://ojs.aaai.org/index.php/AAAI/article/view/5953}
}

@InProceedings{weihs2021advisor,
  title     = {Bridging the Imitation Gap by Adaptive Insubordination},
  author    = {Luca Weihs and Unnat Jain and Iou-Jen Liu and Jordi Salvador and Svetlana Lazebnik and Aniruddha Kembhavi and Alexander Schwing},
  booktitle = {Advances in Neural Information Processing Systems},
  volume    = {34},
  year      = {2021},
  url       = {https://proceedings.neurips.cc/paper/2021/hash/9fc664916bce863561527f06a96f5ff3-Abstract.html}
}

@InProceedings{bahl2023affordances,
  title     = {Affordances from Human Videos as a Versatile Representation for Robotics},
  author    = {Shikhar Bahl and Russell Mendonca and Lili Chen and Unnat Jain and Deepak Pathak},
  booktitle = {Proceedings of the IEEE/CVF Conference on Computer Vision and Pattern Recognition (CVPR)},
  pages     = {13778--13790},
  year      = {2023},
  month     = {June},
  url       = {https://openaccess.thecvf.com/content/CVPR2023/html/Bahl_Affordances_From_Human_Videos_as_a_Versatile_Representation_for_Robotics_CVPR_2023_paper.html}
}

@Misc{sha2026copilot,
  title         = {Efficient and Reliable Teleoperation through Real-to-Sim-to-Real Shared Autonomy},
  author        = {Shuo Sha and Yixuan Wang and Binghao Huang and Antonio Loquercio and Yunzhu Li},
  year          = {2026},
  eprint        = {2603.17016},
  archivePrefix = {arXiv},
  primaryClass  = {cs.RO},
  url           = {https://arxiv.org/abs/2603.17016}
}

@Misc{fu2026capx,
  title         = {{CaP-X}: A Framework for Benchmarking and Improving Coding Agents for Robot Manipulation},
  author        = {Letian Fu and Justin Yu and Karim El-Refai and Ethan Kou and Haoru Xue and Huang Huang and Wenli Xiao and Guanzhi Wang and Fei-Fei Li and Guanya Shi and Jiajun Wu and S. Shankar Sastry and Yuke Zhu and Ken Goldberg and Linxi {Jim} Fan},
  year          = {2026},
  eprint        = {2603.22435},
  archivePrefix = {arXiv},
  primaryClass  = {cs.RO},
  url           = {https://arxiv.org/abs/2603.22435}
}

@Misc{lu2026aspire,
  title         = {{ASPIRE}: Agentic /Skills Discovery for Robotics},
  author        = {Runyu Lu and Yubo Wu and Ethan Kou and Letian Fu and Wenli Xiao and Ajay Mandlekar and Yinzhen Xu and Guanya Shi and Ken Goldberg and Ang Chen and Mosharaf Chowdhury and Yuke Zhu and Linxi {Jim} Fan and Guanzhi Wang},
  year          = {2026},
  eprint        = {2607.00272},
  archivePrefix = {arXiv},
  primaryClass  = {cs.RO},
  url           = {https://arxiv.org/abs/2607.00272}
}

@Misc{xiao2026enpire,
  title         = {{ENPIRE}: Agentic Robot Policy Self-Improvement in the Real World},
  author        = {Wenli Xiao and Jia Xie and Tonghe Zhang and Haotian Lin and Letian Fu and Haoru Xue and Jalen Lu and Yi Yang and Cunxi Dai and Zi Wang and Jimmy Wu and Guanzhi Wang and S. Shankar Sastry and Ken Goldberg and Linxi {Jim} Fan and Yuke Zhu and Guanya Shi},
  year          = {2026},
  eprint        = {2606.19980},
  archivePrefix = {arXiv},
  primaryClass  = {cs.RO},
  url           = {https://arxiv.org/abs/2606.19980}
}

@Misc{hu2026via,
  title         = {{VIA}: Visual Interface Agent for Robot Control},
  author        = {Hengyuan Hu and Priya Sundaresan and Jensen Gao and Dorsa Sadigh},
  year          = {2026},
  eprint        = {2607.11119},
  archivePrefix = {arXiv},
  primaryClass  = {cs.RO},
  url           = {https://arxiv.org/abs/2607.11119}
}

@InProceedings{ichter2023saycan,
  title     = {Do As I Can, Not As I Say: Grounding Language in Robotic Affordances},
  author    = {Brian Ichter and Anthony Brohan and Yevgen Chebotar and Chelsea Finn and Karol Hausman and Alexander Herzog and others},
  booktitle = {Proceedings of The 6th Conference on Robot Learning},
  pages     = {287--318},
  year      = {2023},
  volume    = {205},
  series    = {Proceedings of Machine Learning Research},
  publisher = {PMLR},
  url       = {https://proceedings.mlr.press/v205/ichter23a.html}
}

@InProceedings{liang2023codeaspolicies,
  title     = {Code as Policies: Language Model Programs for Embodied Control},
  author    = {Jacky Liang and Wenlong Huang and Fei Xia and Peng Xu and Karol Hausman and Brian Ichter and Pete Florence and Andy Zeng},
  booktitle = {2023 IEEE International Conference on Robotics and Automation (ICRA)},
  pages     = {9493--9500},
  year      = {2023},
  doi       = {10.1109/ICRA48891.2023.10160591},
  url       = {https://code-as-policies.github.io/}
}

@InProceedings{huang2023voxposer,
  title     = {{VoxPoser}: Composable 3D Value Maps for Robotic Manipulation with Language Models},
  author    = {Wenlong Huang and Chen Wang and Ruohan Zhang and Yunzhu Li and Jiajun Wu and Li Fei-Fei},
  booktitle = {Proceedings of The 7th Conference on Robot Learning},
  pages     = {540--562},
  year      = {2023},
  volume    = {229},
  series    = {Proceedings of Machine Learning Research},
  publisher = {PMLR},
  url       = {https://proceedings.mlr.press/v229/huang23b.html}
}

@Misc{qiu2024wildlma,
  title         = {{WildLMa}: Long Horizon Loco-Manipulation in the Wild},
  author        = {Ri-Zhao Qiu and Yuchen Song and Xuanbin Peng and Sai Aneesh Suryadevara and Ge Yang and Minghuan Liu and Mazeyu Ji and Chengzhe Jia and Ruihan Yang and Xueyan Zou and Xiaolong Wang},
  year          = {2024},
  eprint        = {2411.15131},
  archivePrefix = {arXiv},
  primaryClass  = {cs.RO},
  doi           = {10.48550/arXiv.2411.15131},
  url           = {https://arxiv.org/abs/2411.15131}
}

@InProceedings{ma2024eureka,
  title     = {{Eureka}: Human-Level Reward Design via Coding Large Language Models},
  author    = {Yecheng Jason Ma and William Liang and Guanzhi Wang and De-An Huang and Osbert Bastani and Dinesh Jayaraman and Yuke Zhu and Linxi Fan and Anima Anandkumar},
  booktitle = {International Conference on Learning Representations},
  year      = {2024},
  url       = {https://proceedings.iclr.cc/paper_files/paper/2024/hash/70c26937fbf3d4600b69a129031b66ec-Abstract-Conference.html}
}

@InProceedings{ma2024dreureka,
  title     = {{DrEureka}: Language Model Guided Sim-To-Real Transfer},
  author    = {Yecheng Jason Ma and William Liang and Hung-Ju Wang and Sam Wang and Yuke Zhu and Linxi Fan and Osbert Bastani and Dinesh Jayaraman},
  booktitle = {Robotics: Science and Systems},
  year      = {2024},
  url       = {https://eureka-research.github.io/dr-eureka/}
}

@InProceedings{wang2024rlvlmf,
  title     = {{RL}-{VLM}-F: Reinforcement Learning from Vision Language Foundation Model Feedback},
  author    = {Yufei Wang and Zhanyi Sun and Jesse Zhang and Zhou Xian and Erdem Biyik and David Held and Zackory Erickson},
  booktitle = {Proceedings of the 41st International Conference on Machine Learning},
  pages     = {51484--51501},
  year      = {2024},
  volume    = {235},
  series    = {Proceedings of Machine Learning Research},
  publisher = {PMLR},
  url       = {https://proceedings.mlr.press/v235/wang24bn.html}
}

@InProceedings{dass2023pato,
  title     = {{PATO}: Policy Assisted TeleOperation for Scalable Robot Data Collection},
  author    = {Shivin Dass and Karl Pertsch and Hejia Zhang and Youngwoon Lee and Joseph J. Lim and Stefanos Nikolaidis},
  booktitle = {Proceedings of Robotics: Science and Systems},
  year      = {2023},
  address   = {Daegu, Republic of Korea},
  month     = {July},
  doi       = {10.15607/RSS.2023.XIX.013},
  url       = {https://roboticsproceedings.org/rss19/p013.html}
}

@inproceedings{liu2022robot,
    title = {Robot Learning on the Job: Human-in-the-Loop Autonomy and Learning During Deployment},
    author = {Huihan Liu and Soroush Nasiriany and Lance Zhang and Zhiyao Bao and Yuke Zhu},
    booktitle = {Robotics: Science and Systems (RSS)},
    year = {2023}
}

@misc{i2rt_yam_ultra,
  author = {{I2RT Robotics}},
  title = {YAM Ultra 6-DoF Robotic Arm},
  url = {https://i2rt.com/products/yam-ultra-6-dof-arm}
}

@misc{quest_pro,
  author = {{Meta}},
  title = {Meta Quest Pro},
  url = {https://www.meta.com/quest/quest-pro/}
}


\clearpage
\appendix
\section*{Appendix}
\noindent In this appendix, we include:
\begin{itemize}
    \setlength{\itemsep}{0pt}
    \setlength{\parskip}{0pt}
    \setlength{\topsep}{0.25em}
    \item Appendix~\ref{app:robot-setup}: Robot system setup.
    \item Appendix~\ref{app:guardrail-design}: Emergent guardrails from \methodname.
    \item Appendix~\ref{app:implementation}: Prompts used for
    \methodname{} guardrails.
    \item Appendix~\ref{app:data-feedback-iterations}: Iterative self-refining of \methodname.
    \item Appendix~\ref{app:final-guardrail}: \methodname{} implementation details.
\end{itemize}

We recommend readers visit our project website,
\url{http://guardrail-policy.github.io/}, for additional video results.

\definecolor{guardtriggeraccent}{RGB}{132,45,78}
\newcommand{\guardtrigger}[1]{\par\textcolor{guardtriggeraccent}{\textbf{Trigger:} #1}}
\newcolumntype{L}[1]{>{\raggedright\arraybackslash}p{#1}}
\definecolor{guarddiffadd}{RGB}{231,248,235}
\definecolor{guarddiffdel}{RGB}{253,235,233}
\definecolor{guarddiffnote}{RGB}{0,150,148}
\lstdefinestyle{appendixprompt}{
  style=promptblock,
  basicstyle=\fontencoding{T1}\fontfamily{lmtt}\fontsize{8.5pt}{9pt}\selectfont,
  aboveskip=0.12em,
  belowskip=0.45em
}
\lstdefinestyle{guardcodedel}{
  style=appendixprompt,
  basicstyle=\ttfamily\tiny,
  keepspaces=true,
  backgroundcolor=\color{guarddiffdel}
}
\lstdefinestyle{guardcodeadd}{
  style=appendixprompt,
  basicstyle=\ttfamily\tiny,
  keepspaces=true,
  escapeinside={(*@}{@*)},
  backgroundcolor=\color{guarddiffadd}
}
\newcommand{\guarddiffline}[4]{%
  \begingroup
  \setlength{\fboxsep}{2pt}%
  \noindent\colorbox{#1}{%
    \parbox{\dimexpr\linewidth-2\fboxsep\relax}{%
      {\ttfamily\scriptsize #2\detokenize{#3}}%
      \hfill{\normalfont\sffamily\scriptsize\bfseries\textcolor{guarddiffnote}{#4}}%
    }%
  }%
  \endgroup\par\vspace{0.05em}%
}
\newcommand{\diffadd}[2]{\guarddiffline{guarddiffadd}{+ }{#1}{#2}}
\newcommand{\diffdel}[2]{\guarddiffline{guarddiffdel}{- }{#1}{#2}}
\newcommand{\diffblocktitle}[1]{\par\smallskip\noindent\textbf{\footnotesize #1}\par\vspace{0.05em}}
\newcommand{\diffctx}[1]{\noindent{\ttfamily\scriptsize\detokenize{#1}}\par\vspace{0.03em}}
\newcommand{\guardrailentry}[2]{\leavevmode\smash{\hypertarget{#1}{}}#2}
\newcommand{\guardraildiffref}[3]{Guardrails: \hyperlink{#1}{#2} (Table~\ref{#3})}
\newsavebox{\guarddiffchunkbox}
\newenvironment{guarddiffchunk}[2]{%
  \par\smallskip\noindent
  \begingroup
  \setlength{\fboxsep}{3pt}%
  \begin{lrbox}{\guarddiffchunkbox}%
  \begin{minipage}{\dimexpr\linewidth-2\fboxsep-2\fboxrule\relax}%
  {\footnotesize\textbf{#1}}\hfill{\normalfont\sffamily\scriptsize\bfseries\textcolor{guarddiffnote}{#2}}%
  \par\vspace{0.05em}%
}{%
  \end{minipage}%
  \end{lrbox}%
  \fbox{\usebox{\guarddiffchunkbox}}%
  \endgroup
  \par\smallskip
}
\newcommand{\codeid}[1]{\texttt{\detokenize{#1}}}
\newcommand{\codesnippet}[5]{%
  \begingroup
  \ttfamily\tiny
  \setlength{\parindent}{0pt}%
  \hyperlink{#1}{\textcolor{blue!70!black}{\bfseries\detokenize{#2}}}%
  \detokenize{(}\detokenize{#3}\detokenize{): }\detokenize{#5}\detokenize{; return }\detokenize{#4}%
  \endgroup
}
\newcommand{\manualcodesnippet}[4]{%
  \begingroup
  \ttfamily\tiny
  \setlength{\parindent}{0pt}%
  {\bfseries\detokenize{#1}}%
  \detokenize{(}\detokenize{#2}\detokenize{): }\detokenize{#4}\detokenize{; return }\detokenize{#3}%
  \endgroup
}
\definecolor{guardcodeblockbg}{RGB}{248,248,248}
\definecolor{guardcodeblockrule}{RGB}{205,205,205}
\newcommand{\guardcodeblock}[1]{%
  \begingroup
  \setlength{\fboxsep}{3pt}%
  \setlength{\fboxrule}{0.25pt}%
  \noindent\fcolorbox{guardcodeblockrule}{guardcodeblockbg}{%
    \parbox{\dimexpr\linewidth-2\fboxsep-2\fboxrule\relax}{%
      \setlength{\parindent}{0pt}%
      \setlength{\parskip}{0.12em}%
      #1%
    }%
  }%
  \endgroup
}
\newcommand{\guardcodes}[1]{\guardcodeblock{#1}}
\newcommand{\guardcodesrow}[1]{\multicolumn{2}{L{0.94\linewidth}}{\guardcodes{#1}}\\}
\newcommand{\guardsectionrow}[2]{\multicolumn{2}{>{\raggedright\arraybackslash}p{0.94\linewidth}}{{\bfseries #2 (#1)}}\\*\midrule}
\newcommand{\guardsectionbreak}{\addlinespace[0.65em]\midrule\addlinespace[0.25em]}
\newcommand{\guardrowspace}{\addlinespace[0.25em]}
\newcommand{\guardtablesetup}{%
  \setlength{\tabcolsep}{2.8pt}%
  \setlength{\LTcapwidth}{\textwidth}%
  \renewcommand{\arraystretch}{1.06}%
}

\section{Robot System Setup}
\label{app:robot-setup}

\begin{figure}[ht]
	\centering
	\includegraphics[width=0.8\linewidth]{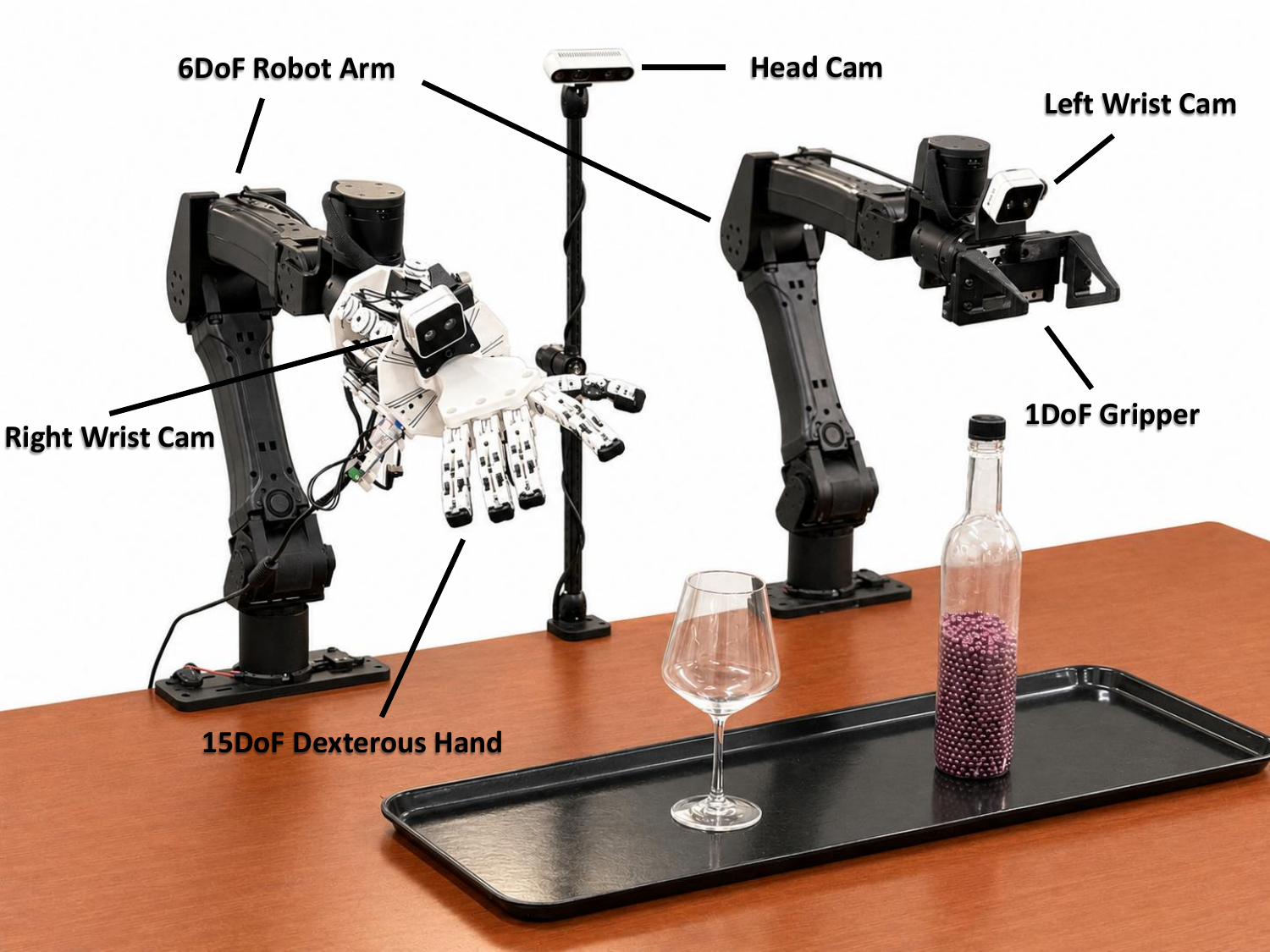}
	\caption{Robot system setup for dexterous tasks.}
	\label{fig:teleop-setup}
\end{figure}

We use the bimanual I2RT YAM Ultra~\citep{i2rt_yam_ultra} as our robot platform. Each arm provides
6 DoFs for arm motion and an additional DoF for the gripper. For dexterous tasks, we
replace the right-arm gripper with the CRAFT hand~\citep{lin2026craft}, which
provides 15 DoFs for more dexterous manipulation. Teleoperation demonstrations are
collected with a Meta Quest Pro VR headset~\citep{quest_pro}. Gripper-based teleoperation uses handheld
controllers for precise arm and gripper motion, while dexterous teleoperation
uses VR hand tracking to capture hand movements directly. We
follow OpenTV~\citep{cheng2024opentelevision} to track hand motions relative to
the headset, and the hand-tracking stream provides calibrated commands for the
dexterous hand. Our dexterous robot system setup is shown in
\figref{fig:teleop-setup}.

\section{Emergent Guardrails from \methodname}
\label{app:guardrail-design}

\begin{figure}[ht]
	\centering
	\includegraphics[width=\linewidth]{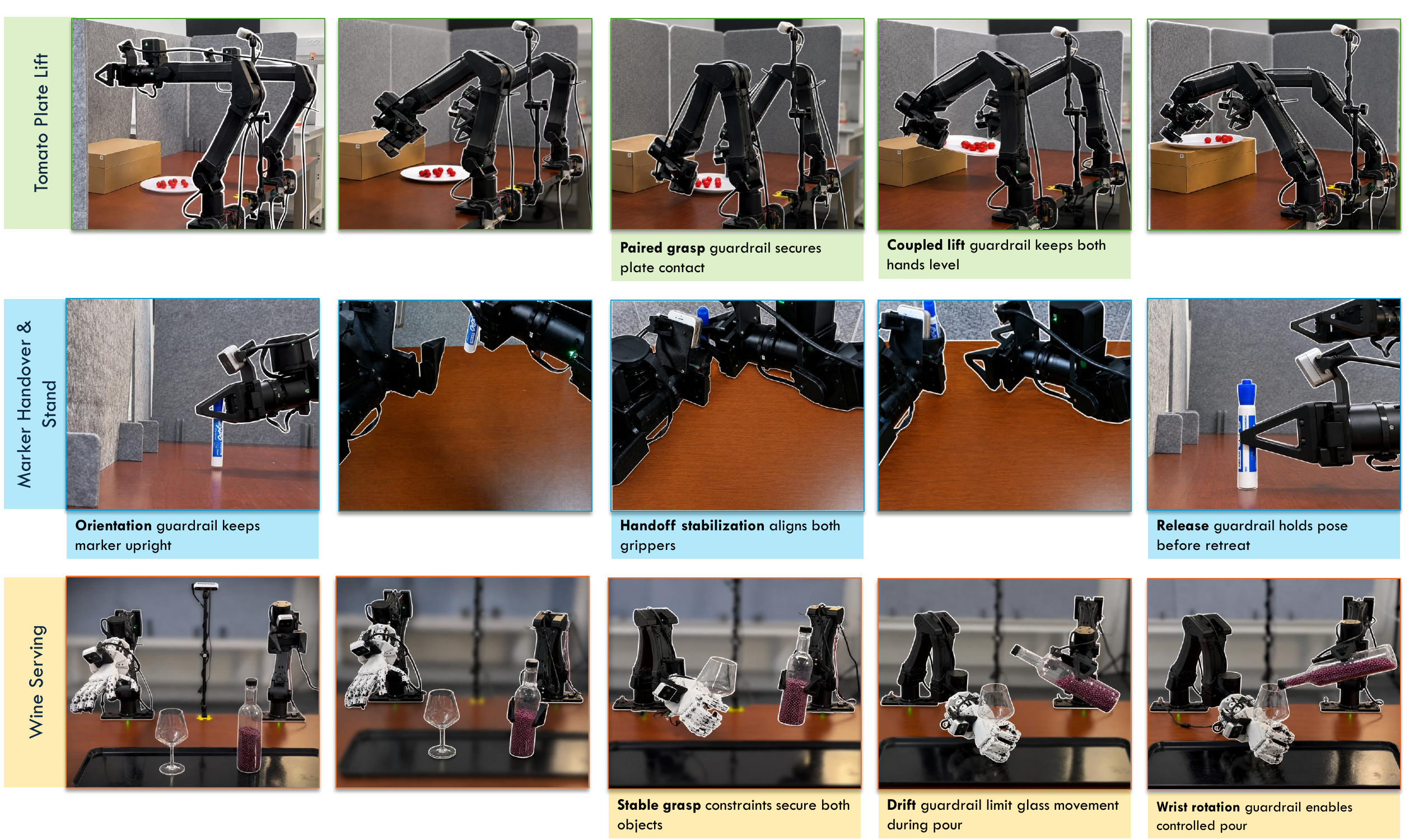}
	\caption{\textbf{Phase-wise \methodname{} guardrail progression.}
	Across the three evaluation tasks, trajectory feedback refines the initial
	guardrails into the optimal \Gopt{}: coordinated plate lifting and carry
	stabilization, marker handoff and release stabilization, and bottle-glass
	alignment and rotation assistance for controlled pouring. Task descriptions can be found at \href{https://guardrail-policy.github.io/\#overview}{the project webpage}.}
	\label{fig:guardrail-progress}
	\vspace{-0.5em}
\end{figure}

For the guardrails comparisons in this section, \Gsetglide{} denotes the set of
individual restrictions implemented by \Gopt{}, while \Gsetde{} denotes the
set implemented by \Gde{}. The
\methodname{} guardrails \Gopt{} and domain-expert hardcoded guardrails \Gde{} are built \textbf{separately}:
\Gopt{} does not build upon, inherit from, or refine \Gde{}. Rows labeled $\Gsetglide{} \setminus \Gsetde{}$ denote emergent guardrails
from \methodname{} that go beyond domain-expert hardcoded guardrails, and rows labeled
$\Gsetglide{} \cap \Gsetde{}$ denote restrictions shared by both methods. Rows labeled
$\Gsetde{} \setminus \Gsetglide{}$ denote domain-expert hardcoded guardrails that are
absent from \methodname{}.
\figref{fig:guardrail-progress} shows how the \methodname{} guardrails evolve over each task,
and \tabref{tab:plate-guardrail-restrictions},
\tabref{tab:marker-guardrail-restrictions}, and
\tabref{tab:wine-guardrail-restrictions} summarize the details of \Gsetglide{} and \Gsetde{}
restrictions.

\paragraph{\tomatoemoji{}~\emph{Tomato plate transfer}.}
\tabref{tab:plate-guardrail-restrictions} summarizes the details of \Gsetglide{} and \Gsetde{} for \emph{Tomato plate transfer} task.
\textbf{\methodname{} guardrails.} \Gopt{} rate-limits end-effector motion and prevents large downward moves from
the teleop start pose. It enters carry mode when both triggers are pressed or
both grippers are at least 35\% closed. In carry mode, the script treats the
end effectors as a coupled carrier: it slows vertical motion, limits
acceleration, prevents a large post-grasp drop, keeps the two sides level,
preserves grasp width, blocks inward compression, and strongly resists wrist
tilt. It also changes gripper control so both grippers toggle together through a
two-controller chord, with delayed reopen and rate-limited gripper commands to ensure gentle opening and closing.

\textbf{Domain-expert hardcoded guardrails.} \Gde{} covers a subset of the above restrictions: after both
grippers close, it keeps the plate nearly level, keeps the grasp width near the
reference width at grasp, and limits wrist rotation away from the grasp pose.

\begingroup
\guardtablesetup
\begin{longtable}{L{0.29\linewidth}L{0.65\linewidth}}
		\caption{\tomatoemoji{}~\textbf{\emph{Tomato Plate Transfer}: interpreting the guardrails discovered by \methodname{}.}
			Collecting stable demonstrations for bimanual plate transport is hard, as human operators struggle to keep the plate level, the grasp secure, and the motion smooth simultaneously.
			\methodname{} discovers guardrails that match domain-expert hardcoded ones, while new task-specific constraints emerge from task descriptions and recorded trajectories, without human-written failure diagnoses.}
		\label{tab:plate-guardrail-restrictions}\\
	\toprule
	\textbf{Guardrails} & \textbf{Implementation details} \\
	\midrule
	\endfirsthead
		\caption[]{\tomatoemoji{}~\textbf{Tomato plate transfer guardrails} (continued).}\\
	\textbf{Guardrails} & \textbf{Implementation details} \\
	\midrule
	\endhead
	\multicolumn{2}{r}{Continued on next page}\\
	\endfoot
	\bottomrule
	\endlastfoot
						\guardsectionrow{$\Gsetglide{} \setminus \Gsetde{}$}{Emergent \methodname{} guardrails beyond expert hardcoding}
							\guardrailentry{guardrail:plate-workspace-motion}{Bounded workspace and stabilized pre-grasp approach} &
						While approaching the plate (before the grippers close), EEF speed is bounded ($\leq 0.26$\,m/s) to prevent abrupt pre-grasp motion.\par To prevent the EEFs from contacting the base surface, downward displacement from the episode's initial height is bounded ($\leq 0.35$\,m). \guardtrigger{During pre-grasp approach, before carry stabilization is active.} \\
						\guardcodesrow{\codesnippet{code:plate-filter-position}{_filter_position}{raw_pos, prev_pos, ...}{filtered_pos}{p = ema(raw_pos, prev_pos); step = clip_speed(p - prev_pos); step = clip_accel(step, prev_step); filtered_pos = prev_pos + step}\par\codesnippet{code:plate-filter-single}{filter_single}{target_pose, init_pose, ...}{filtered_pose}{pose.z = max(target_pose.z, init_pose.z - down_margin); pose.pos = _filter_position(...); pose.rot = _filter_rotation(...); filtered_pose = pose}}
						\guardrowspace
						\guardrailentry{guardrail:plate-bimanual-motion}{Plate carried as a coupled system with bounded midpoint motion} &
				Once both grippers are sufficiently closed ($\geq$35\%), the two EEFs are treated as a coupled system governed through their shared midpoint.\newline
				To keep transport smooth and prevent spills, midpoint speed and acceleration are bounded separately in the vertical (speed $\leq$0.055\,m/s, acceleration $\leq$0.18\,m/s$^2$) and horizontal (speed $\leq$0.24\,m/s, acceleration $\leq$0.85\,m/s$^2$) directions; each EEF speed is also individually bounded ($\leq$0.22\,m/s).\newline
				To prevent the plate from descending back toward the surface mid-transport, the midpoint is bounded from dropping more than 0.03\,m below its height at grasp. \guardtrigger{When both grippers $\geq$35\% closed, or when both controller triggers are pressed.} \\
						\guardcodesrow{\codesnippet{code:plate-filter-carry-midpoint}{_filter_carry_midpoint}{raw_mid, prev_mid}{filtered_mid}{step.xy = clip_xy(raw_mid.xy - prev_mid.xy); step.z = clip_z(raw_mid.z - prev_mid.z); filtered_mid = prev_mid + step}\par\codesnippet{code:plate-filter-pair}{filter_pair}{left_target_pose, right_target_pose, ...}{left_filtered, right_filtered}{mid = _filter_carry_midpoint((left.pos + right.pos)/2, prev_mid); half = clip_carry_halfsep(left.pos - right.pos); left_filtered.pos = mid + half; right_filtered.pos = mid - half}}
						\guardrowspace
						\guardrailentry{guardrail:plate-coupled-gripper}{Grippers act as a synchronized pair with confirmed grasp before release} &
					Both grippers open and close in unison: a simultaneous press of both VR controller triggers is required, preventing asymmetric grip on the plate.\newline
					To ensure the grasp is secure before any release, reopening is locked until both grippers have remained $\geq$90\% closed for at least 1.5\,s.\newline
					To prevent sudden grip changes that could jolt the plate, gripper command speed is bounded ($\leq$0.8\,travel/s). \guardtrigger{On a two-controller trigger chord; reopening only after both grippers satisfy the 90\% closed, 1.5-s hold condition.} \\
						\guardcodesrow{\codesnippet{code:plate-toggle-grippers}{toggle_plate_grippers_from_triggers}{left_arm, right_arm, ...}{None}{target = open if both_closed else close; if target == open and hold_time < min_hold: None; set(left, right, target)}\par\codesnippet{code:plate-limit-gripper-rate}{limit_plate_gripper_command_rate}{arm_state, dt, ...}{filtered_pos}{target = min(target, max_close); filtered_pos = clip(target, prev - max_speed * dt, prev + max_speed * dt)}}
						\guardsectionbreak
						\guardsectionrow{$\Gsetglide{} \cap \Gsetde{}$}{Shared \methodname{}/domain-expert hardcoded guardrails}
						\guardrailentry{guardrail:plate-height-sync}{Grippers kept at matching heights to avoid side-to-side tilt} &
				To keep the plate from tilting sideways, the height difference between the left and right grippers is bounded ($\leq$0.8\,cm under \Gsetglide{}, $\leq$0.5\,cm under \Gsetde{}). \\
						\guardcodesrow{\codesnippet{code:plate-filter-pair}{filter_pair}{left_target_pose, right_target_pose, ...}{left_filtered, right_filtered}{half.z = clip(half.z, -0.5 * max_height_diff, 0.5 * max_height_diff); poses satisfy dz <= 0.008}}
						\guardrowspace
				\guardrailentry{guardrail:plate-grasp-width}{Gripper separation locked to preserve the grasp on the plate} &
			To prevent the plate from slipping out of the grasp, the distance between the two grippers is kept close to what it was when the plate was first picked up ($\leq$2\,cm deviation under \Gsetglide{}, $\leq$1\,cm under \Gsetde{}); \Gsetglide{} also prevents the grippers from squeezing closer together, which could dislodge the plate. \\
						\guardcodesrow{\codesnippet{code:plate-carry-reference}{_ensure_carry_reference}{left_target_pose, right_target_pose}{None}{ref_width = norm(left.pos - right.pos); ref_rot = left.rot, right.rot}\par\codesnippet{code:plate-filter-pair}{filter_pair}{left_target_pose, right_target_pose, ...}{left_filtered, right_filtered}{half = clip_to_ref_width(raw_half, ref_half); dsep <= 0.02; compression = 0}}
						\guardrowspace
				\guardrailentry{guardrail:plate-wrist-orientation}{Gripper angles anchored to the initial grasp to prevent forward/backward tip and plate bending} &
			To prevent the plate from tipping forward or backward during transport, each gripper's angle is steered back toward the orientation it had at the moment of grasping. \Gsetglide{} does this gradually, while \Gsetde{} enforces a strict limit on how far each gripper can rotate ($\leq$0.1\,rad). \\
						\guardcodesrow{\codesnippet{code:plate-filter-rotation}{_filter_rotation}{init_rot, target_rot, weight}{filtered_rot}{filtered_rot = project_rotation((1 - weight) * target_rot + weight * init_rot); carry_weight = 0.9}}
\end{longtable}
\endgroup

\paragraph{\markeremoji{}~\emph{Marker handover \& stand}.}
\tabref{tab:marker-guardrail-restrictions} summarizes the details of \Gsetglide{} and \Gsetde{} for \emph{Marker handover \& stand} task.
\textbf{\methodname{} guardrails.}
\Gopt{}
rate-limits target motion, optionally keeps targets above a predefined table
height, maintains horizontal spacing between grippers, and detects transfer when
the two grippers are near each other and either both triggers are pressed or a
gripper is holding the marker. During transfer, it damps relative motion and
aligns gripper heights. It also reduces roll/pitch tilt during normal motion,
more strongly while carrying and during release. At the release stage, when a gripper opens from a
firm grasp, the filter holds the release pose briefly, then allows a small
upward retreat while tightly limiting sideways motion.

\textbf{Domain-expert hardcoded guardrails.}
\Gde{} overlaps on upright orientation: it blocks
roll and pitch while yaw and translation remain active.

\begingroup
\guardtablesetup
\begin{longtable}{L{0.29\linewidth}L{0.65\linewidth}}
		\caption{\markeremoji{}~\textbf{\emph{Marker Handover \& Stand}: interpreting the guardrails discovered by \methodname{}.}
			Collecting stable demonstrations for marker handover and upright placement is hard, as human operators struggle to keep the marker vertical, coordinate the transfer, and release without tipping.
			\methodname{} discovers guardrails that match domain-expert hardcoded ones, while new task-specific constraints emerge from task descriptions and recorded trajectories, without human-written failure diagnoses.}
		\label{tab:marker-guardrail-restrictions}\\
	\toprule
	\textbf{Guardrails} & \textbf{Implementation details} \\
	\midrule
	\endfirsthead
		\caption[]{\markeremoji{}~\textbf{Marker handover \& stand guardrails} (continued).}\\
	\textbf{Guardrails} & \textbf{Implementation details} \\
	\midrule
	\endhead
	\multicolumn{2}{r}{Continued on next page}\\
	\endfoot
	\bottomrule
	\endlastfoot
						\guardsectionrow{$\Gsetglide{} \setminus \Gsetde{}$}{Emergent \methodname{} guardrails beyond expert hardcoding}
							\guardrailentry{guardrail:marker-motion-smoothing}{EEF motion smoothed to avoid knocking over the marker} &
					To avoid abrupt motions that can knock over the upright marker, each EEF's motion is smoothed and bounded: total speed is capped ($\leq$0.42\,m/s), with directional caps in the horizontal plane ($\leq$0.45\,m/s) and vertical direction ($\leq$0.20\,m/s). Total acceleration is also bounded ($\leq$2.50\,m/s$^2$), with separate horizontal ($\leq$3.00\,m/s$^2$) and vertical ($\leq$1.40\,m/s$^2$) limits. \guardtrigger{Throughout marker pickup, handover, and placement.} \\
						\guardcodesrow{\codesnippet{code:marker-limit-step}{_limit_step}{raw_step, prev_step}{step}{step = clip_speed_xy_z(raw_step); step = prev_step + clip_accel_xy_z(step - prev_step)}\par\codesnippet{code:marker-filter-position}{_filter_position}{raw_pos, prev_pos, prev_step, ...}{filtered_pos, step}{raw_pos = _apply_table_guard(raw_pos); target = prev_pos + alpha * (raw_pos - prev_pos); step = _limit_step(target - prev_pos, prev_step); filtered_pos = prev_pos + step}}
						\guardrowspace
				\guardrailentry{guardrail:marker-gripper-spacing}{Table clearance and gripper spacing prevent scraping and collisions} &
				To keep the EEFs from scraping the table during marker pickup and placement, targets are kept at least 1.5\,cm above the table when a table height is configured.\newline To prevent the grippers from colliding with each other outside the transfer stage, the two grippers are kept at least 5.0\,cm apart horizontally. \guardtrigger{Throughout marker pickup, handoff and placement.} \\
						\guardcodesrow{\codesnippet{code:marker-table-guard}{_apply_table_guard}{pos}{guarded}{guarded = pos; guarded.z = max(pos.z, table_z + clearance)}\par\codesnippet{code:marker-enforce-min-distance}{_enforce_min_xy_distance}{left_pos, right_pos, ...}{left_pos, right_pos}{if dist_xy(left_pos, right_pos) < min_distance: left_pos.xy += correction; right_pos.xy -= correction}}
						\guardrowspace
				\guardrailentry{guardrail:marker-handover-stabilization}{Close-range handoff damping keeps the receiver from overshooting the marker} &
			When the two grippers move close enough to pass the marker between them (within 18\,cm horizontally and 10\,cm vertically), the filter treats the motion as a handoff attempt once the operator signals transfer intent with both controller triggers. During this transfer, the horizontal spacing limit relaxes from 5.0\,cm to 3.5\,cm so the grippers can meet near the marker. The filter also slows sudden relative motion between the grippers, preventing the receiving gripper from overshooting the marker, and bounds the vertical mismatch between grippers ($\leq$4.0\,cm). \guardtrigger{During a nearby handoff attempt.} \\
						\guardcodesrow{\codesnippet{code:marker-shape-handoff}{_shape_handoff_targets}{left_pos, right_pos, ...}{left_pos, right_pos}{mid = (left_pos + right_pos)/2; half = prev_half + gain * (raw_half - prev_half); half.z = clip(half.z); left_pos = mid + half; right_pos = mid - half}\par\codesnippet{code:marker-should-handoff}{should_stabilize_pen_handoff}{left_arm, right_arm, ...}{active}{near = xy_dist <= xy_window and z_dist <= z_window; active = near and (both_triggers or left_holding or right_holding)}}
						\guardrowspace
						\guardrailentry{guardrail:marker-release-stabilization}{EEf release pose held steady so the marker can stand before EEF retreat} &
					To prevent the marker from tipping at the final placement, opening a previously firm grasp first freezes the release pose for 0.55\,s. Once the gripper has opened past 50\%, the EEF is allowed to lift by 1.5\,cm; lateral retreat is only allowed after the gripper opens past 20\%, and is bounded to a 0.8\,cm radius at 1.5\,cm/s. Upward speed is capped at 0.12\,m/s, and gripper opening speed is capped at 0.90 travel/s. \guardtrigger{When the marker-holding gripper starts opening from at least 70\% closed.} \\
							\guardcodesrow{\codesnippet{code:marker-release-filter}{_apply_release_filter}{side, target_pose, close_fraction}{filtered}{if opening_from_grasp: ref = target_pose; if hold: pose = ref; else: pose.xy = clip_radius(target.xy, ref.xy); pose.z = ref.z + lift; filtered = pose}\par\codesnippet{code:marker-limit-gripper-rate}{limit_pen_gripper_command_rate}{arm_state, dt, ...}{filtered_pos}{if opening and delay_remaining > 0: target = prev; filtered_pos = clip(target, prev - speed * dt, prev + speed * dt)}}
							\guardsectionbreak
							\guardsectionrow{$\Gsetglide{} \cap \Gsetde{}$}{Shared \methodname{}/domain-expert hardcoded guardrails}
							\guardrailentry{guardrail:marker-keep-upright}{Yaw-only wrist orientation keeps the marker vertical} &
					To avoid rolling or pitching the marker over, \Gsetglide{} steers each wrist toward a yaw-only orientation while still allowing operator-controlled yaw. \Gsetde{} implements the same safety intent more directly by blocking roll/pitch commands, removing operator-commanded wrist rotations that would tilt the marker forward/backward or sideways.. \\
							\guardcodesrow{\codesnippet{code:marker-filter-rotation}{_filter_rotation_weight}{init_rot, target_rot, weight}{filtered_rot}{yaw_target = yaw_only(init_rot, target_rot); blend = (1 - weight) * target_rot + weight * yaw_target; filtered_rot = project_rotation(blend)}}
\end{longtable}
\endgroup

\paragraph{\wineemoji{}~\emph{Wine serving}.}
\tabref{tab:wine-guardrail-restrictions} summarizes the details of \Gsetglide{} and \Gsetde{} for \emph{Wine serving} task.
\textbf{\methodname{} guardrails.} \Gopt{} keeps arm
targets inside the robot workspace, rate-limits target changes, separates the
arms, and bounds finger travel. After both
objects are grasped, it smooths gripper and dexterous hand commands and bounds glass and
bottle tilt so the grasps remain stable during carry and pour. When wrist
rotation indicates pour intent, it assists bottle-mouth alignment and can use
height and lateral error to gate larger bottle tilt.

\textbf{Domain-expert hardcoded guardrails.}
\Gde{} arbitrates simultaneous arm commands, keeping only one arm move at a time, and map wrist-roll/pinch
gestures to pouring and gripper toggling, without enforcing grasp stability,
object drift, or bottle-glass alignment.

\begingroup
\guardtablesetup
\begin{longtable}{L{0.29\linewidth}L{0.65\linewidth}}
		\caption{\wineemoji{}~\textbf{\emph{Wine Serving}: interpreting the guardrails discovered by \methodname{}.}
			Collecting stable demonstrations for wine serving is hard: operators must keep both grasps stable, align the bottle with the cup, and pour without spilling.
			\methodname{} discovers task-specific guardrails from task descriptions and recorded trajectories, without human-written failure diagnoses.}
		\label{tab:wine-guardrail-restrictions}\\
	\toprule
	\textbf{Guardrails} & \textbf{Implementation details} \\
	\midrule
	\endfirsthead
		\caption[]{\wineemoji{}~\textbf{Wine serving guardrails} (continued).}\\
	\textbf{Guardrails} & \textbf{Implementation details} \\
	\midrule
	\endhead
	\multicolumn{2}{r}{Continued on next page}\\
	\endfoot
	\bottomrule
	\endlastfoot
						\guardsectionrow{$\Gsetglide{} \setminus \Gsetde{}$}{Emergent \methodname{} guardrails beyond expert hardcoding}
								\guardrailentry{guardrail:wine-workspace-speed}{Workspace and speed limits keep bottle and cup in a stable operating region} &
						To keep both manipulated objects in a stable operating region, each arm's EEF height is constrained to $[0.085, 0.72]$\,m. Commanded motion is rate-limited so translation changes by at most 0.32\,m/s and rotation by at most 2.60\,rad/s. \guardtrigger{Throughout.} \\
							\guardcodesrow{\codesnippet{code:wine-clamp-pose}{clamp_pose}{pose}{pose, changed}{clamped = clip(pose.xyz, lower_bound, upper_bound); changed = clamped != pose.xyz; pose.xyz = clamped}\par\codesnippet{code:wine-rate-limit}{_rate_limit}{side, pose, dt, ...}{pose}{pose.pos = limit_translation(prev.pos, pose.pos, max_speed * dt); pose.rot = limit_rotation(prev.rot, pose.rot, max_angular * dt)}\par\codesnippet{code:wine-task-defaults}{_apply_task_defaults}{args, base_argv, ...}{None}{if flag not in base_argv: args.flag = guarded_default}}
						\guardrowspace
						\guardrailentry{guardrail:wine-safety-distance-finger}{Hand separation and finger limits prevent collision and cup over-grip} &
					To prevent the bottle and cup end effectors from colliding, the two EEFs are maintained at least 10.5\,cm apart. To avoid over-gripping or destabilizing the cup, thumb bend, finger side-splay, and grip are bounded to 58\%, 22\%, and 68\% of their respective motion ranges. \guardtrigger{Throughout.} \\
							\guardcodesrow{\codesnippet{code:wine-separate-arms}{_separate_arms}{left, right, ...}{left, right}{if dist(left, right) < min_dist: center = (left.pos + right.pos)/2; left.pos = center + offset; right.pos = center - offset}\par\codesnippet{code:wine-guard-craft-targets}{guard_craft_targets}{targets, ...}{guarded_targets}{for motor, raw in targets: guarded_targets[motor] = clip_by_motor_type(motor, raw)}}
						\guardrowspace
							\guardrailentry{guardrail:wine-cup-bottle-tilt}{Carry and aligned-pour tilt limits keep the cup upright while pouring} &
					To allow pickup and setup while still preventing spills, the cup can tilt up to 85$^\circ$ during carry and the bottle up to 90$^\circ$ before pouring. A bottle tilt of at least 36$^\circ$ is treated as pour intent. Alignment is then checked from EEF-derived tool points: the bottle-mouth point and cup-rim point must be within 13\,cm laterally, with the bottle mouth 6--44\,cm above the rim. Under this aligned-pour condition, the cup is tightened to a near-upright limit ($\leq$18$^\circ$) while the bottle is allowed to tilt farther ($\leq$135$^\circ$) for pouring. \guardtrigger{After both objects are grasped; stricter cup stabilization applies during aligned pouring.} \\
							\guardcodesrow{\codesnippet{code:wine-pose-axis}{_pose_axis_closest_to_up}{pose}{axis}{axis = max(local_axes, key=lambda axis: dot(pose.rot @ axis, WORLD_UP))}\par\codesnippet{code:wine-allowed-bottle-tilt}{_allowed_bottle_tilt}{left_pose, right_pose, ...}{max_tilt}{pour_requested = tilt(left_pose, bottle_axis) >= pour_start_tilt; max_tilt = carry_tilt; if aligned and height_ok: max_tilt = pour_tilt}}
						\guardrowspace
							\guardrailentry{guardrail:wine-pouring-assistance}{Tilt boost and mouth guidance help the bottle reach the cup} &
							When the operator starts a pour motion (bottle tilt $>$28$^\circ$), bottle tilt is amplified by 1.85$\times$ so the bottle can reach a useful pouring angle. As the pour becomes clearer (alignment assistance starts at 32$^\circ$), the bottle mouth is guided toward a point 16\,cm above the cup rim, with the correction capped at 18\,cm. \guardtrigger{When wrist rotation indicates pouring intent.} \\
							\guardcodesrow{\codesnippet{code:wine-boost-bottle-pour-tilt}{_boost_bottle_pour_tilt}{left_pose, ...}{left_pose}{if tilt > boost_start: tilt = min(boost_start + gain * (tilt - boost_start), pour_max_tilt); left_pose = set_tilt(left_pose, tilt)}\par\codesnippet{code:wine-assist-alignment}{_assist_alignment}{left_pose, right_pose, ...}{left_pose}{error = cup_rim + target_height - bottle_mouth; left_pose.pos += strength * clip_norm(error, max_correction)}}
							\guardsectionbreak
							\guardsectionrow{$\Gsetde{} \setminus \Gsetglide{}$}{Domain-expert hardcoded guardrails absent from \methodname{}}
								\guardrailentry{guardrail:wine-one-arm}{Dominant-arm arbitration keeps only one EEF move at a time} &
							\Gsetde{} keeps only one EEF move at a time by selecting the arm with the larger motion score at each timestep. Once an arm is selected, the other arm must become meaningfully more active before control switches, preventing frame-to-frame toggling. The non-selected arm command is discarded and its retargeter is synchronized to the current target so it does not accumulate stale motion. \guardtrigger{Throughout, when both arms are active.} \\
							\guardcodesrow{\manualcodesnippet{choose_guardrail_side}{left_arm, right_arm, previous_side, ...}{selected_side, left_score, right_score}{left_score = motion_score(left_arm) + toggle_bonus; right_score = motion_score(right_arm) + toggle_bonus; selected_side = argmax_with_hysteresis(left_score, right_score)}\par\manualcodesnippet{sync_retargeter_side_to_arm}{retargeter, side, arm_state}{None}{state = retargeter.states[side]; state.filtered_target_pose = arm_state.target_pose; state.last_good_target_pose = arm_state.target_pose}}
						\guardrowspace
							\guardrailentry{guardrail:wine-translation-lock-toggle}{Roll-to-pour locks bottle position; pinch toggles grippers for easier control} &
						When the left wrist holding the bottle rolls below $-60^\circ$, \Gsetde{} freezes translation for both arms while preserving roll-only bottle rotation, preventing the bottle from drifting during a pour gesture. Separately, a pinch gesture toggles an available gripper open or closed. \guardtrigger{Translation lock triggers when left wrist roll crosses $-60^\circ$. A gripper toggle triggers when the active hand's pinch signal rises above 0.75 after first relaxing below 0.35, so holding a pinch does not repeatedly toggle.} \\
							\guardcodesrow{\manualcodesnippet{left_roll_translation_lock_state}{left_pose, ...}{roll_lock_active, roll_deg}{roll = relative_roll(reference_pose, left_pose); roll_lock_active = roll <= lock_threshold; roll_deg = roll}\par\manualcodesnippet{freeze_translation_allow_roll}{arm_state, pose}{guarded_pose}{guarded_pose = roll_only_delta(arm_state.target_pose, pose); guarded_pose.pos = arm_state.target_pose.pos}\par\manualcodesnippet{toggle_gripper_from_pinch}{arm_state}{target_label}{target_label = close if gripper_is_open else open; set_gripper(target_label)}}
\end{longtable}
\endgroup
\clearpage

\section{Prompts used for
    \methodname{} Guardrails}
\label{app:implementation}

\noindent\textbf{\tomatoemoji{}~\emph{Tomato plate transfer}.}
\begin{nolinenumbers}
\begin{lstlisting}[style=appendixprompt]
I have a bimanual teleoperation script at <script_name>, where the readings of two controllers from VR headset are mapped to the robot commands through inverse kinematics. I am now collecting data for a new task, that bimanually picking up and lifting a plate full of cherry tomatoes onto a box. Please identify the potential failure cases in the teleoperation process, design constraints and apply the filter that will help to make the human teleoperation easier.
\end{lstlisting}
\end{nolinenumbers}

\noindent\textbf{\markeremoji{}~\emph{Marker handover \& stand}.}
\begin{nolinenumbers}
\begin{lstlisting}[style=appendixprompt]
I have a bimanual teleoperation script at <script_name>, where the readings of two controllers from VR headset are mapped to the robot commands through inverse kinematics. I am now collecting data for a new task, that using one gripper to pick up the upright marker on the table, handing it over to the other gripper, and place it vertically on the table. Please identify the potential failure cases in the teleoperation process, design constraints and apply the filter that will help to make the human teleoperation easier. Please add a flag to enable this filter.
\end{lstlisting}
\end{nolinenumbers}

\noindent\textbf{\wineemoji{}~\emph{Wine serving}.}
\begin{nolinenumbers}
\begin{lstlisting}[style=appendixprompt]
I have a bimanual teleoperation script at <script_name>, where the readings of two controllers from VR headset are mapped to the robot commands through inverse kinematics and retargeting. The left arm is equipped with a gripper, and the right arm is equipped with a dexterous hand. I am now collecting data for a new task, that use left gripper to pick up the wine bottle, and right hand to pick up the wine cup, and after aligning, we pour wine from bottle to the cup, ideally during pouring, the liquid should not spill and finally put both bottle and cup on the table. Please identify the potential failure cases in the teleoperation process, design constraints and apply the filter that will help to make the human teleoperation easier. Write a new script for this.
\end{lstlisting}
\end{nolinenumbers}

\section{Iterative Self-refining of \methodname}
\label{app:data-feedback-iterations}

Each refinement round uses the same trajectory-feedback prompt, with the
dataset path updated to the current round. \methodname{} inspects recorded
head/wrist videos, robot proprioception, and raw/guarded commands to produce
the log diagnoses below and revise the task-specific guardrail code.
The human performs teleoperation and may optionally supply outcome labels or
intuitive feedback; the diagnoses do not require human-written failure
annotations. The shared prompt supports different tasks, while the task and
hardware context, recorded data, and generated guardrails vary.
\par\smallskip
\noindent\textbf{Trajectory-based guardrail refinement.}
\begin{nolinenumbers}
\begin{lstlisting}[style=appendixprompt]
I collected 10 new episodes in <dataset_path>. Please analyze the dataset in detail, identify the failed episodes and failure modes, find the core vulnerability in the current guardrails, and patch the guardrail code if necessary. Look at the teleoperation readings, robot actions, states, and videos; compare success and failure episodes; and summarize what the previous code changes fixed, what still failed, and what should be kept. Feel free to add or delete filters and constraints instead of only tuning parameters. The goal is to make teleoperation easier, safer, more precise, and more efficient, so the next data-collection round produces more successful trajectories.
\end{lstlisting}
\end{nolinenumbers}

\paragraph{\tomatoemoji{}~\emph{Tomato plate transfer}.}
There are two iterations after
inspecting two 10-episode datasets. In both
updates, the grasp guard was left intact because the episodes reached full
gripper close; the code edits targeted the loaded-carry phase where tomatoes
spilled or the task became too slow.

\diffblocktitle{\texttt{iteration1}: make loaded carry gentler after spill episodes}
{\noindent\footnotesize\textit{Log diagnosis: episodes 6--9 spilled after the
plate was already grasped; failed runs had faster loaded transport and larger
left/right height mismatch.}\par}

\begin{guarddiffchunk}{\guardraildiffref{guardrail:plate-height-sync}{EEF height synchronization}{tab:plate-guardrail-restrictions}}{Tightening shared stability bound}
\diffdel{    parser.add_argument("--plate-carry-max-height-diff", type=float, default=0.012)}{}
\diffadd{    parser.add_argument("--plate-carry-max-height-diff", type=float, default=0.008)}{smaller height mismatch}
\end{guarddiffchunk}

\begin{guarddiffchunk}{\guardraildiffref{guardrail:plate-bimanual-motion}{Bimanual motion constraints}{tab:plate-guardrail-restrictions}}{Changing speed limits; adding acceleration guard}
\diffdel{    parser.add_argument("--plate-filter-alpha", type=float, default=0.45)}{}
\diffadd{    parser.add_argument("--plate-filter-alpha", type=float, default=0.40)}{gentler loaded update}
\diffdel{    parser.add_argument("--plate-carry-max-ee-speed", type=float, default=0.18)}{}
\diffadd{    parser.add_argument("--plate-carry-max-ee-speed", type=float, default=0.14)}{slower loaded transport}
\diffadd{    parser.add_argument("--plate-carry-max-ee-accel", type=float, default=0.45)}{new acceleration guard}
\diffdel{        left_pos = self._filter_position(left_pos, left_prev, self.config.carry_max_speed)}{}
\diffdel{        right_pos = self._filter_position(right_pos, right_prev, self.config.carry_max_speed)}{}
\diffadd{        left_pos = self._filter_position(..., self.prev_left_step, self.config.carry_max_accel)}{}
\diffadd{        right_pos = self._filter_position(..., self.prev_right_step, self.config.carry_max_accel)}{}
\end{guarddiffchunk}

\diffblocktitle{\texttt{iteration2}: make transport faster without relaxing vertical safety}
{\noindent\footnotesize\textit{Log diagnosis: episode 7 fully closed but spilled
during carry; the successful episodes were slow because one isotropic filter
throttled horizontal and vertical motion together.}\par}
\begin{guarddiffchunk}{\guardraildiffref{guardrail:plate-bimanual-motion}{Bimanual motion constraints}{tab:plate-guardrail-restrictions}}{Move the plate as one object}
\diffdel{        left_pos = self._filter_position(..., self.config.carry_max_speed, self.config.carry_max_accel)}{}
\diffdel{        right_pos = self._filter_position(..., self.config.carry_max_speed, self.config.carry_max_accel)}{}
\diffadd{    carry_max_xy_speed: float}{separate horizontal and vertical caps}
\diffadd{    carry_max_z_speed: float}{}
\diffadd{    carry_max_xy_accel: float}{}
\diffadd{    carry_max_z_accel: float}{}
\diffadd{    def _filter_carry_midpoint(self, raw_mid, prev_mid):}{}
\diffadd{        step[:2] = clamp_vector_norm(step[:2], max_xy_step)}{faster table-to-box travel}
\diffadd{        step[2] = float(np.clip(step[2], -max_z_step, max_z_step))}{slow lift/drop motion}
\diffadd{        filtered_mid = self._filter_carry_midpoint(raw_mid, prev_mid)}{}
\diffadd{        left_pos = filtered_mid + filtered_half_sep}{restore grasp offsets}
\diffadd{        right_pos = filtered_mid - filtered_half_sep}{}
\diffdel{    parser.add_argument("--plate-carry-max-ee-speed", type=float, default=0.14)}{}
\diffadd{    parser.add_argument("--plate-carry-max-ee-speed", type=float, default=0.22)}{final per-arm safety cap}
\diffadd{    parser.add_argument("--plate-carry-max-xy-speed", type=float, default=0.24)}{}
\diffadd{    parser.add_argument("--plate-carry-max-z-speed", type=float, default=0.055)}{}
\end{guarddiffchunk}

\begin{guarddiffchunk}{\guardraildiffref{guardrail:plate-wrist-orientation}{Wrist orientation restraint}{tab:plate-guardrail-restrictions}}{Relaxing an over-strict orientation lock}
\diffdel{    parser.add_argument("--plate-carry-orientation-weight", type=float, default=1.0)}{}
\diffadd{    parser.add_argument("--plate-carry-orientation-weight", type=float, default=0.9)}{small tilt allowed}
\end{guarddiffchunk}

In short, \texttt{iteration1} slowed down the plate motion, and
\texttt{iteration2} made the loaded plate behave less like two independent
hands and more like a cautious tray.

\paragraph{\markeremoji{}~\emph{Marker handover \&\ stand}.}
The first iteration removed the guardrail-induced lag, then
added an explicit release phase. The second iteration made that final release
slower and better supported.

\diffblocktitle{\texttt{iteration1}: remove lag, then guard the release}
{\noindent\footnotesize\textit{Log diagnosis: in the dataset, raw controller
motion reached p95 about \(0.23\,m/s\), while commanded EE speed was only about
\(0.12\,m/s\), with vertical commands pinned by the old \(0.08\,m/s\) cap. In the failure episodes, the marker got upright briefly, then tipped or fell during final opening and retract.}\par}

\begin{guarddiffchunk}{\guardraildiffref{guardrail:marker-handover-stabilization}{Handover stabilization}{tab:marker-guardrail-restrictions}}{Relaxing over-damped guidance}
\diffdel{    parser.add_argument("--pen-orientation-weight", type=float, default=0.85)}{}
\diffadd{    parser.add_argument("--pen-orientation-weight", type=float, default=0.15)}{operator can aim}
\diffdel{    parser.add_argument("--pen-handoff-orientation-weight", type=float, default=0.95)}{}
\diffadd{    parser.add_argument("--pen-handoff-orientation-weight", type=float, default=0.35)}{guided but movable}
\diffdel{    parser.add_argument("--pen-min-ee-xy-distance", type=float, default=0.045)}{}
\diffadd{    parser.add_argument("--pen-handoff-min-ee-xy-distance", type=float, default=0.035)}{handoff specific}
\end{guarddiffchunk}

\begin{guarddiffchunk}{\guardraildiffref{guardrail:marker-motion-smoothing}{Motion smoothing}{tab:marker-guardrail-restrictions}}{Changing speed damping parameter to remove lag}
\diffdel{    parser.add_argument("--pen-filter-alpha", type=float, default=0.35)}{}
\diffadd{    parser.add_argument("--pen-filter-alpha", type=float, default=0.70)}{follow operator faster}
\diffdel{    parser.add_argument("--pen-max-ee-speed", type=float, default=0.22)}{}
\diffadd{    parser.add_argument("--pen-max-ee-speed", type=float, default=0.42)}{}
\diffdel{    parser.add_argument("--pen-max-z-speed", type=float, default=0.08)}{}
\diffadd{    parser.add_argument("--pen-max-z-speed", type=float, default=0.20)}{faster vertical motion}
\diffadd{    parser.add_argument("--pen-max-xy-speed", type=float, default=0.45)}{slower horizontal motion}
\diffdel{    parser.add_argument("--pen-max-ee-accel", type=float, default=0.70)}{}
\diffadd{    parser.add_argument("--pen-max-ee-accel", type=float, default=2.50)}{catch up quickly}
\diffadd{    parser.add_argument("--pen-max-xy-accel", type=float, default=3.00)}{}
\diffadd{    parser.add_argument("--pen-max-z-accel", type=float, default=1.40)}{}
\end{guarddiffchunk}

\begin{guarddiffchunk}{\guardraildiffref{guardrail:marker-release-stabilization}{Gripper release stabilization}{tab:marker-guardrail-restrictions}}{Adding new release guard}
\diffdel{    parser.add_argument("--pen-gripper-max-speed", type=float, default=0.80)}{}
\diffadd{    parser.add_argument("--pen-gripper-max-speed", type=float, default=1.20)}{softer release speed}
\diffadd{    release_start_close_fraction: float}{}
\diffadd{    release_end_close_fraction: float}{}
\diffadd{    release_hold_s: float}{}
\diffadd{    release_xy_radius: float}{}
\diffadd{    release_orientation_weight: float}{}
\diffadd{        self.release_state = {"left": self._new_release_state(), "right": self._new_release_state()}}{}
\diffadd{    def _apply_release_filter(...):}{freeze upright pose before retreat}
\diffadd{        state["ref_pos"] = target_pose[:3, 3].copy()}{}
\diffadd{        state["ref_rot"] = target_pose[:3, :3].copy()}{}
\diffadd{        desired_xy = ref_pos[:2] + clamp_vector_norm(...)}{tiny lateral motion}
\diffadd{        release_z = ref_pos[2] + release_lift_height * lift_progress}{small upward retreat}
\end{guarddiffchunk}

\diffblocktitle{\texttt{iteration2}: delay and soften the final finger opening}
{\noindent\footnotesize\textit{Log diagnosis: the failures were no longer mainly lag,
but topples during or immediately after final release.}\par}
\begin{guarddiffchunk}{\guardraildiffref{guardrail:marker-release-stabilization}{Gripper release stabilization}{tab:marker-guardrail-restrictions}}{Changing release functional form}
\diffdel{    parser.add_argument("--pen-carry-orientation-weight", type=float, default=0.55)}{}
\diffadd{    parser.add_argument("--pen-carry-orientation-weight", type=float, default=0.45)}{lighter carry damping}
\diffadd{    parser.add_argument("--pen-release-lift-start-close-fraction", type=float, default=0.50)}{lift later}
\diffadd{    parser.add_argument("--pen-release-lift-height", type=float, default=0.015)}{small guided lift}
\diffdel{        if elapsed >= self.config.release_retreat_s and close_fraction <= end_close:}{wrong timing}
\diffadd{        if close_fraction <= end_close: state["open_frames"] += 1}{track true open time}
\diffadd{        if state["open_frames"] * self.dt >= self.config.release_retreat_s:}{hold after open}
\diffadd{    parser.add_argument("--pen-gripper-max-open-speed", type=float, default=0.90)}{opening gets own cap}
\diffadd{    parser.add_argument("--pen-gripper-release-delay-s", type=float, default=0.15)}{short release pause}
\diffadd{    parser.add_argument("--pen-gripper-release-delay-close-fraction", type=float, default=0.65)}{}
\end{guarddiffchunk}

In short, \texttt{iteration1} first made the marker follow the operator, then
turned release into a guarded manipulation phase. \texttt{iteration2} made the
last opening slower and better supported. The idea is simple: once the
marker is upright, preserve that pose for a few more ticks while the gripper
open.

\paragraph{\wineemoji{}~\emph{Wine serving}.}
The first iteration learned the true tool frame, then stopped
blocking acquisition motion. The second added pour authority and bottle-to-glass
alignment help once the operator clearly tries to pour.

\diffblocktitle{\texttt{iteration1}: learn the tool frame and remove setup caps}
{\noindent\footnotesize\textit{Log diagnosis: in the dataset,
both end effectors looked upward at startup because the guardrails assumed local
\(z\) was upright; FK showed local \(y\) was actually closest to world up. Meanwhile, the glass capped exactly at \(12^\circ\) and the
bottle capped exactly at \(28^\circ\), so the guardrails were blocking necessary
wrist rotation before pouring.}\par}
\begin{guarddiffchunk}{\guardraildiffref{guardrail:wine-cup-bottle-tilt}{Cup/Bottle tilt (angle from global upward z axis) constrains}{tab:wine-guardrail-restrictions}}{Learn the true tool frame}
\diffdel{def _axis_vector(axis: str) -> np.ndarray:}{fixed tool frame}
\diffadd{def _axis_vector(axis: str) -> np.ndarray | None:}{support auto axis}
\diffadd{    if axis == "auto": return None}{}
\diffadd{def _pose_axis_closest_to_up(pose: np.ndarray) -> np.ndarray:}{learn true upright axis}
\diffadd{    return max(valid, key=lambda axis: float(np.dot(rot @ axis, WORLD_UP))).copy()}{}
\diffdel{    cup_upright_axis: np.ndarray = field(default_factory=lambda: _axis_vector("z"))}{}
\diffdel{    bottle_upright_axis: np.ndarray = field(default_factory=lambda: _axis_vector("z"))}{}
\diffadd{    cup_upright_axis: np.ndarray | None = None}{auto-learn cup frame}
\diffadd{    bottle_upright_axis: np.ndarray | None = None}{auto-learn bottle frame}
\diffdel{    cup_upright_max_rad: float = math.radians(12.0)}{}
\diffadd{    cup_upright_max_rad: float = math.radians(18.0)}{strict only near pour}
\diffadd{    cup_carry_max_tilt_rad: float = math.radians(85.0)}{free cup pickup}
\diffdel{    bottle_carry_max_tilt_rad: float = math.radians(28.0)}{}
\diffadd{    bottle_carry_max_tilt_rad: float = math.radians(90.0)}{free bottle setup}
\diffdel{            right, changed = _cap_tilt(right, self._upright_axis("right"), self.config.cup_upright_max_rad)}{}
\diffadd{            cup_max_tilt = self.config.cup_carry_max_tilt_rad}{phase-aware cup cap}
\diffadd{            if pour_requested and pour_aligned: cup_max_tilt = self.config.cup_upright_max_rad}{}
\end{guarddiffchunk}

\begin{guarddiffchunk}{\guardraildiffref{guardrail:wine-workspace-speed}{Workspace \& speed limits}{tab:wine-guardrail-restrictions}}{Restoring setup wrist authority}
\diffdel{        "rotation_alpha": (0.50, "--rotation-alpha")}{}
\diffadd{        "rotation_alpha": (0.75, "--rotation-alpha")}{more wrist response}
\diffdel{        "max_target_angular_speed": (1.15, "--max-target-angular-speed")}{}
\diffadd{        "max_target_angular_speed": (2.80, "--max-target-angular-speed")}{faster target rotation}
\diffdel{    parser.add_argument("--guardrail-max-angular-speed", type=float, default=1.05)}{}
\diffadd{    parser.add_argument("--guardrail-max-angular-speed", type=float, default=2.60)}{faster guarded wrist}
\end{guarddiffchunk}

\diffblocktitle{\texttt{iteration2}: add pour authority and alignment assistance}
{\noindent\footnotesize\textit{Log diagnosis: in the dataset,
the operator wrist reached about \(103^\circ\), but the robot bottle tilt
stopped around \(88^\circ\), so the robot still could not pour.}\par}
\begin{guarddiffchunk}{\guardraildiffref{guardrail:wine-cup-bottle-tilt}{Cup/Bottle tilt (angle from global upward z axis) constrains}{tab:wine-guardrail-restrictions}}{Increasing pour authority}
\diffdel{    bottle_pour_max_tilt_rad: float = math.radians(125.0)}{}
\diffadd{    bottle_pour_max_tilt_rad: float = math.radians(135.0)}{higher pour ceiling}
\diffadd{    bottle_pour_tilt_boost_gain: float = 1.85}{amplify pour intent}
\diffadd{    bottle_pour_tilt_boost_start_rad: float = math.radians(28.0)}{}
\diffdel{        "max_arm_joint_step": (0.026, "--max-arm-joint-step")}{}
\diffadd{        "max_arm_joint_step": (0.045, "--max-arm-joint-step")}{allow pour motion}
\end{guarddiffchunk}

\begin{guarddiffchunk}{\guardraildiffref{guardrail:wine-pouring-assistance}{Pouring assistance}{tab:wine-guardrail-restrictions}}{Adding pouring assistance}
\diffadd{    align_assist: bool = True}{turn on assistance}
\diffadd{    align_assist_start_rad: float = math.radians(32.0)}{assist after tilt}
\diffadd{    align_assist_strength: float = 0.65}{soft pull}
\diffadd{    align_assist_target_height_m: float = 0.16}{aim above glass}
\diffadd{    align_assist_max_correction_m: float = 0.18}{cap the help}
\diffadd{            left = self._boost_bottle_pour_tilt(left, reasons)}{boost before gates}
\diffadd{            left = self._assist_alignment(left, right, reasons)}{pull toward glass}
\diffadd{    def _boost_bottle_pour_tilt(...):}{scale operator tilt}
\diffadd{        boosted_tilt = start + (tilt - start) * gain}{}
\diffadd{    def _assist_alignment(...):}{soft bottle-to-glass correction}
\diffadd{        desired_bottle_point = cup_point + np.asarray(...)}{}
\diffadd{        out[:3, 3] = out[:3, 3] + amount * correction}{}
\end{guarddiffchunk}

In short, \texttt{iteration1} fixed the frame the guardrails used and made
constraints phase-aware instead of blocking pickup. \texttt{iteration2} added
assistance once the operator clearly tried to pour.

\section{\methodname{} Implementation Details}
\label{app:final-guardrail}

\paragraph{\tomatoemoji{}~\emph{Tomato plate transfer}.}
\diffblocktitle{Naive VR teleoperation}
\begin{nolinenumbers}
\begin{lstlisting}[style=guardcodedel]
# Raw VR teleoperation: each arm follows its controller independently.
left_target_pose = compute_target_pose(...); left_q = ik(left_target_pose)
right_target_pose = compute_target_pose(...); right_q = ik(right_target_pose)

# Raw grippers: each trigger toggles only its own gripper.
left_trigger -> toggle_gripper_from_trigger(left_arm)
right_trigger -> toggle_gripper_from_trigger(right_arm)
\end{lstlisting}
\end{nolinenumbers}

\diffblocktitle{\methodname{} implementation}
\begin{nolinenumbers}
\begin{lstlisting}[style=guardcodeadd]
class PlateLiftTeleopFilter:
    # Task-space filter inserted between raw VR target poses and IK.

    def __init__(self, config, frequency):
        self.config = config
        self.dt = 1.0 / max(float(frequency), 1e-6)
        self.prev_left_pos = self.prev_right_pos = None
        self.prev_mid_step = None
        self.carry_active = False
        self.carry_left_pos = self.carry_right_pos = None
        self.carry_left_rot = self.carry_right_rot = None

    def reset(self):
        # Function: clear stale carry state when teleop pauses or reinitializes.
        ...

(*@\hypertarget{code:plate-filter-position}{}@*)    def _filter_position(self, raw_pos, prev_pos, max_speed=None, prev_step=None, max_accel=None):
        # Function: light single-arm smoothing for approach.
        ...

(*@\hypertarget{code:plate-filter-carry-midpoint}{}@*)    def _filter_carry_midpoint(self, raw_mid, prev_mid):
        # Function: move the shared plate center. XY can move faster for
        # transport, while Z is deliberately slower to avoid drops and spills.
        filtered = prev_mid + self.config.pos_alpha * (raw_mid - prev_mid)
        step = filtered - prev_mid
        step[:2] = clamp_vector_norm(step[:2], carry_max_xy_speed * dt)
        step[2] = clip(step[2], -carry_max_z_speed * dt, carry_max_z_speed * dt)
        step_delta = step - self.prev_mid_step
        step_delta[:2] = clamp_vector_norm(step_delta[:2], carry_max_xy_accel * dt * dt)
        step_delta[2] = clip(step_delta[2], -carry_max_z_accel * dt * dt, carry_max_z_accel * dt * dt)
        self.prev_mid_step = step
        return prev_mid + step

(*@\hypertarget{code:plate-filter-rotation}{}@*)    def _filter_rotation(self, init_rot, target_rot, weight=None):
        # Function: blend wrist rotation back toward the carry reference so
        # the loaded plate remains close to level.
        return project_rotation((1.0 - weight) * target_rot + weight * init_rot)

(*@\hypertarget{code:plate-filter-single}{}@*)    def filter_single(self, side, target_pose, init_pose):
        # Function: before paired carry, smooth one arm and block large
        # downward moves below the reference pose.
        raw_pos[2] = max(raw_pos[2], init_pose[2, 3] - down_margin)
        filtered[:3, :3] = self._filter_rotation(init_rot, target_rot)
        return filtered

(*@\hypertarget{code:plate-carry-reference}{}@*)    def _ensure_carry_reference(self, left_target_pose, right_target_pose):
        # Function: on carry entry, remember grasp width, height balance, and
        # wrist orientation as the reference for stable transport.
        self.carry_left_pos = prev_left_pos or left_target_pose[:3, 3]
        self.carry_right_pos = prev_right_pos or right_target_pose[:3, 3]
        self.carry_left_rot = left_target_pose[:3, :3]
        self.carry_right_rot = right_target_pose[:3, :3]

(*@\hypertarget{code:plate-filter-pair}{}@*)    def filter_pair(self, left_target_pose, right_target_pose, left_init_pose, right_init_pose, stabilize=False):
        # Function: main guardrails.  If not carrying, fall back to single-arm
        # smoothing.  If carrying, command both arms as one plate support.
        if not stabilize:
            return filter_single(left), filter_single(right)
        self._ensure_carry_reference(left_target_pose, right_target_pose)
        init_mid = 0.5 * (carry_left_pos + carry_right_pos)
        init_half_sep = 0.5 * (carry_left_pos - carry_right_pos)
        raw_mid = 0.5 * (raw_left_pos + raw_right_pos)
        raw_half_sep = 0.5 * (raw_left_pos - raw_right_pos)
        raw_mid[2] = max(raw_mid[2], init_mid[2] - carry_down_margin)
        half_delta = carry_differential_gain * (raw_half_sep - init_half_sep)
        half_delta[:2] = clamp_vector_norm(half_delta[:2], 0.5 * carry_max_separation_delta)
        half_delta[2] = clip(half_delta[2], -0.5 * carry_max_height_diff, 0.5 * carry_max_height_diff)
        min_xy_norm = init_xy_norm - 0.5 * carry_max_compression
        filtered_mid = self._filter_carry_midpoint(raw_mid, prev_mid)
        left_pos = filtered_mid + filtered_half_sep
        right_pos = filtered_mid - filtered_half_sep
        left_rot = self._filter_rotation(carry_left_rot, left_target_rot, carry_orientation_weight)
        right_rot = self._filter_rotation(carry_right_rot, right_target_rot, carry_orientation_weight)
        return left_filtered, right_filtered


def should_stabilize_plate(left_arm, right_arm, left_state, right_state, args):
    # Function: phase detector; enter paired carry when both triggers are
    # pressed or both grippers are closed enough.
    return both_triggers or (left_closed and right_closed)


(*@\hypertarget{code:plate-toggle-grippers}{}@*)def toggle_plate_grippers_from_triggers(left_arm, right_arm, args):
    # Function: paired gripper toggle; two-hand trigger intent opens/closes
    # both grippers together, with a minimum hold before reopening.
    target_label = "open" if min(left_closed, right_closed) >= reopen_threshold else "close"
    if target_label == "open" and now - last_close_time < reopen_min_hold_time:
        return None
    set_gripper_target_label(left_arm, target_label)
    set_gripper_target_label(right_arm, target_label)


(*@\hypertarget{code:plate-limit-gripper-rate}{}@*)def limit_plate_gripper_command_rate(arm_state, args, dt, command_updated=True):
    # Function: gripper guard; clip over-close and limit per-step gripper
    # motion for gentler grasp and release.
    target_pos = clip_to_plate_gripper_max_close_fraction(target_pos)
    filtered_pos = clip(target_pos, prev_pos - max_speed * dt, prev_pos + max_speed * dt)
\end{lstlisting}
\end{nolinenumbers}

\paragraph{\markeremoji{}~\emph{Marker handover \& stand}.}
\diffblocktitle{Naive VR teleoperation}
\begin{nolinenumbers}
\begin{lstlisting}[style=guardcodedel]
# Raw VR teleoperation: both arms follow controller poses directly.
left_target_pose = compute_target_pose(...); left_q = ik(left_target_pose)
right_target_pose = compute_target_pose(...); right_q = ik(right_target_pose)

# Raw grippers: gripper targets follow trigger/toggle commands immediately.
update_gripper_from_controller(left_arm, left_state, gripper_mode, ...)
update_gripper_from_controller(right_arm, right_state, gripper_mode, ...)
\end{lstlisting}
\end{nolinenumbers}

\diffblocktitle{\methodname{} implementation}
\begin{nolinenumbers}
\begin{lstlisting}[style=guardcodeadd]
class PenHandoverTeleopFilter:
    # Function: task-space guardrails for pickup, handoff, carry, and vertical release.

    def __init__(self, config, frequency):
        self.config = config
        self.dt = 1.0 / max(float(frequency), 1e-6)
        self.prev_left_pos = self.prev_right_pos = None
        self.prev_left_step = self.prev_right_step = None
        self.prev_close_fraction = {"left": None, "right": None}
        self.release_state = {"left": self._new_release_state(), "right": self._new_release_state()}

    def reset(self):
        # Function: clear smoothed positions and release states when teleop resets.
        ...

    def _new_release_state(self):
        # Function: create one side's release-phase memory.
        return {"active": False, "frames": 0, "open_frames": 0, "ref_pos": None, "ref_rot": None, "pos": None}

(*@\hypertarget{code:marker-table-guard}{}@*)    def _apply_table_guard(self, pos):
        # Function: optional table-height guard, preventing the target from dipping below the table plus clearance.
        guarded = pos.copy()
        guarded[2] = max(guarded[2], table_z + table_clearance)
        return guarded

(*@\hypertarget{code:marker-limit-step}{}@*)    def _limit_step(self, raw_step, prev_step):
        # Function: smooth EE motion with separate XY/Z speed and acceleration caps.
        step[:2] = clamp_vector_norm(step[:2], max_xy_speed * dt)
        step[2] = clip(step[2], -max_z_speed * dt, max_z_speed * dt)
        step_delta[:2] = clamp_vector_norm(step_delta[:2], max_xy_accel * dt * dt)
        step_delta[2] = clip(step_delta[2], -max_z_accel * dt * dt, max_z_accel * dt * dt)
        return step

(*@\hypertarget{code:marker-filter-position}{}@*)    def _filter_position(self, raw_pos, prev_pos, prev_step):
        # Function: apply table guard, low-pass filtering, and step limiting.
        raw_pos = self._apply_table_guard(raw_pos)
        filtered = prev_pos + pos_alpha * (raw_pos - prev_pos)
        step = self._limit_step(filtered - prev_pos, prev_step)
        return self._apply_table_guard(prev_pos + step), step

(*@\hypertarget{code:marker-filter-rotation}{}@*)    def _filter_rotation_weight(self, init_rot, target_rot, weight):
        # Function: damp roll/pitch by blending target rotation toward a yaw-only target.
        yaw_target = yaw_only_target_rotation(init_rot, target_rot)
        return project_rotation((1.0 - weight) * target_rot + weight * yaw_target)

(*@\hypertarget{code:marker-release-filter}{}@*)    def _apply_release_filter(self, side, target_pose, close_fraction):
        # Function: the key final-placement guardrails.  When fingers begin
        # opening after a firm grasp, anchor pose/rotation, prevent side shoves,
        # and allow a slow upward retreat only after a hold.
        opening = prev_close >= release_start_close_fraction and close_fraction < prev_close - 0.02
        if opening and not state["active"]:
            state["active"] = True
            state["ref_pos"] = target_pose[:3, 3].copy()
            state["ref_rot"] = target_pose[:3, :3].copy()
            state["pos"] = target_pose[:3, 3].copy()
        if not lift_allowed:
            filtered_pos = ref_pos.copy()
        else:
            desired_xy = ref_pos[:2] + clamp_vector_norm(target_pose[:2, 3] - ref_pos[:2], release_xy_radius)
            release_z = ref_pos[2] + release_lift_height * lift_progress
            filtered_pos[:2] = prev_pos[:2] + clamp_vector_norm(desired_xy - prev_pos[:2], release_max_xy_speed * dt)
            filtered_pos[2] = prev_pos[2] + clip(release_z - prev_pos[2], 0.0, release_max_up_speed * dt)
        filtered[:3, 3] = filtered_pos
        filtered[:3, :3] = project_rotation((1.0 - release_orientation_weight) * target_rot + release_orientation_weight * ref_rot)
        return filtered

(*@\hypertarget{code:marker-enforce-min-distance}{}@*)    def _enforce_min_xy_distance(self, left_pos, right_pos, left_init_pos, right_init_pos, min_distance=None):
        # Function: keep grippers separated so the two arms do not collide or squeeze the marker handoff.
        correction = 0.5 * (min_distance - distance) * direction
        left_pos[:2] += correction
        right_pos[:2] -= correction
        return left_pos, right_pos

(*@\hypertarget{code:marker-shape-handoff}{}@*)    def _shape_handoff_targets(self, left_pos, right_pos):
        # Function: during handoff, damp relative hand motion and limit height mismatch.
        raw_mid = 0.5 * (left_pos + right_pos)
        half_sep = prev_half_sep + handoff_relative_gain * (raw_half_sep - prev_half_sep)
        half_z = clip(0.5 * (left_pos[2] - right_pos[2]), -0.5 * handoff_max_height_diff, 0.5 * handoff_max_height_diff)
        return raw_mid + half_sep, raw_mid - half_sep

    def filter_single(self, side, target_pose, init_pose, close_fraction=None):
        # Function: single-arm carry guard; smooth motion, use stronger orientation damping when holding,
        # and run the release filter when the gripper starts opening.
        filtered_pos, step = self._filter_position(raw_pos, prev_pos, prev_step)
        weight = carry_orientation_weight if close_fraction >= carry_grasp_close_threshold else orientation_weight
        filtered[:3, :3] = self._filter_rotation_weight(init_rot, target_rot, weight)
        return self._apply_release_filter(side, filtered, close_fraction)

    def filter_pair(self, left_target_pose, right_target_pose, left_init_pose, right_init_pose, handoff=False, left_close_fraction=None, right_close_fraction=None):
        # Function: paired marker guard.  Handoff mode shapes the two targets together;
        # all paired motion enforces spacing, orientation damping, and release filtering.
        if handoff:
            left_raw, right_raw = self._shape_handoff_targets(left_raw, right_raw)
        left_pos, left_step = self._filter_position(left_raw, left_prev, prev_left_step)
        right_pos, right_step = self._filter_position(right_raw, right_prev, prev_right_step)
        if handoff:
            left_pos, right_pos = self._shape_handoff_targets(left_pos, right_pos)
        min_distance = handoff_min_ee_xy_distance if handoff else min_ee_xy_distance
        left_pos, right_pos = self._enforce_min_xy_distance(left_pos, right_pos, left_init_pos, right_init_pos, min_distance)
        left_weight = handoff_orientation_weight if handoff else carry_or_default(left_close_fraction)
        right_weight = handoff_orientation_weight if handoff else carry_or_default(right_close_fraction)
        left_filtered = self._apply_release_filter("left", left_filtered, left_close_fraction)
        right_filtered = self._apply_release_filter("right", right_filtered, right_close_fraction)
        return left_filtered, right_filtered


(*@\hypertarget{code:marker-should-handoff}{}@*)def should_stabilize_pen_handoff(left_arm, right_arm, left_target_pose, right_target_pose, left_state, right_state, args):
    # Function: handoff phase detector.  Only activate paired handoff shaping when the grippers are near
    # each other in XY/Z and the operator indicates transfer intent or one gripper is holding the marker.
    near_handoff = xy_distance <= handoff_xy_window and z_distance <= handoff_z_window
    return near_handoff and (both_triggers or left_holding or right_holding)


(*@\hypertarget{code:marker-limit-gripper-rate}{}@*)def limit_pen_gripper_command_rate(arm_state, args, dt, command_updated=True):
    # Function: release timing guard.  Clip over-close, delay first opening after a firm grasp,
    # and use separate open/close speeds so the marker is not kicked over at release.
    target_pos = clip_to_pen_gripper_max_close_fraction(target_pos)
    if opening and delay_remaining > 0.0:
        target_pos = prev_pos
    step_speed = max_open_speed if opening else max_close_speed if closing else max_speed
    filtered_pos = clip(target_pos, prev_pos - step_speed * dt, prev_pos + step_speed * dt)
\end{lstlisting}
\end{nolinenumbers}

\paragraph{\wineemoji{}~\emph{Wine serving}.}
\diffblocktitle{Naive VR teleoperation}
\begin{nolinenumbers}
\begin{lstlisting}[style=guardcodedel]
# Raw arm path: raw Quest retargeting output is commanded directly.
yam_output = retargeter.update(frame, dt=period, now=frame.timestamp)
left_ok = command_pose(left_arm, yam_output.left_pose, max_arm_joint_step, ...)
right_ok = command_pose(right_arm, yam_output.right_pose, max_arm_joint_step, ...)

# Raw CRAFT path: raw hand landmarks become motor targets directly.
craft_action_targets, signals, last_seen = retarget_craft(...)
craft.write_raw(craft_action_targets)
\end{lstlisting}
\end{nolinenumbers}

\diffblocktitle{\methodname{} implementation}
\begin{nolinenumbers}
\begin{lstlisting}[style=guardcodeadd]
(*@\hypertarget{code:wine-pose-axis}{}@*)def _pose_axis_closest_to_up(pose):
    # Function: auto-calibrate which local tool axis is upright at reset.
    # This avoids assuming glass/bottle local z is the upright direction.
    return max([x, -x, y, -y, z, -z], key=lambda axis: dot(pose[:3, :3] @ axis, WORLD_UP))


class GuardrailBounds:
    # Function: workspace clamp for bottle and glass poses.
(*@\hypertarget{code:wine-clamp-pose}{}@*)    def clamp_pose(self, pose):
        pose[x, y, z] = clip_to_bounds(pose[x, y, z])
        return pose, changed


class WinePourGuardrail:
    # Function: post-process raw arm poses before the base teleop loop commands IK.

    def reset(self, left_pose, right_pose):
        # Function: clear previous pose memory and learn bottle/glass upright axes
        # from the calibrated poses when axes are set to auto.
        self.previous.clear()
        if bottle_upright_axis is None:
            self._learned_axes["left"] = _pose_axis_closest_to_up(left_pose)
        if cup_upright_axis is None:
            self._learned_axes["right"] = _pose_axis_closest_to_up(right_pose)

    def _upright_axis(self, side):
        # Function: use configured tool axis when provided, otherwise use the
        # learned axis from reset/calibration.
        return configured_axis or learned_axis or _axis_vector("y")

    def filter_output(self, output, dt):
        # Function: main arm-pose guardrail pipeline.  It preserves the raw loop
        # but edits retargeted poses before command_pose sees them.
        left, right = copy(output.left_pose), copy(output.right_pose)
        reasons = []
        pour_requested = self._pour_requested(left)
        pour_aligned = self._pour_geometry_ok(left, right)
        left = self._boost_bottle_pour_tilt(left, reasons)
        cup_max_tilt = cup_carry_max_tilt
        if lock_cup_during_pour and pour_requested and pour_aligned:
            cup_max_tilt = cup_upright_max_tilt
        right = _cap_tilt(right, self._upright_axis("right"), cup_max_tilt)
        left = _cap_tilt(left, self._upright_axis("left"), self._allowed_bottle_tilt(left, right, reasons))
        left = self._assist_alignment(left, right, reasons)
        left = left_bounds.clamp_pose(left)
        right = right_bounds.clamp_pose(right)
        left, right = self._separate_arms(left, right, reasons)
        left = self._rate_limit("left", left, dt, reasons)
        right = self._rate_limit("right", right, dt, reasons)
        output.left_pose, output.right_pose = left, right
        return output

(*@\hypertarget{code:wine-boost-bottle-pour-tilt}{}@*)    def _boost_bottle_pour_tilt(self, left_pose, reasons):
        # Function: once the operator starts pouring, amplify bottle tilt so the
        # robot can reach a useful pour angle without requiring extreme VR wrist motion.
        tilt = _tilt_angle(left_pose, self._upright_axis("left"))
        if tilt <= bottle_pour_tilt_boost_start:
            return left_pose
        start = bottle_pour_tilt_boost_start
        boosted_tilt = start + (tilt - start) * bottle_pour_tilt_boost_gain
        return self._set_tilt(left_pose, bottle_axis, min(boosted_tilt, bottle_pour_max_tilt))

(*@\hypertarget{code:wine-assist-alignment}{}@*)    def _assist_alignment(self, left_pose, right_pose, reasons):
        # Function: during pour intent, softly pull the bottle mouth toward a
        # point above the glass rim, capped by a maximum correction distance.
        if tilt <= align_assist_start:
            return left_pose
        bottle_point = _tool_point(left_pose, bottle_mouth_offset)
        cup_point = _tool_point(right_pose, cup_rim_offset)
        desired_bottle_point = cup_point + [0.0, 0.0, align_assist_target_height]
        correction = clamp_norm(desired_bottle_point - bottle_point, align_assist_max_correction)
        left_pose[:3, 3] += align_assist_strength * correction
        return left_pose

(*@\hypertarget{code:wine-allowed-bottle-tilt}{}@*)    def _allowed_bottle_tilt(self, left_pose, right_pose, reasons):
        # Function: allow pickup/setup tilt, then optionally require bottle-glass
        # alignment and valid height before permitting large pour tilt.
        if requested_tilt < pour_start_tilt:
            return bottle_carry_max_tilt
        if not require_pour_alignment:
            return bottle_pour_max_tilt
        aligned = lateral_distance(bottle_point, cup_point) <= align_radius
        height_ok = pour_min_height <= bottle_point[2] - cup_point[2] <= pour_max_height
        return bottle_pour_max_tilt if aligned and height_ok else bottle_carry_max_tilt

(*@\hypertarget{code:wine-separate-arms}{}@*)    def _separate_arms(self, left, right, reasons):
        # Function: keep bottle and glass end-effectors separated to avoid collisions.
        if distance >= min_ee_distance:
            return left, right
        center = 0.5 * (left[:3, 3] + right[:3, 3])
        left[:3, 3] = center + 0.5 * min_ee_distance * direction
        right[:3, 3] = center - 0.5 * min_ee_distance * direction
        return left_bounds.clamp_pose(left), right_bounds.clamp_pose(right)

(*@\hypertarget{code:wine-rate-limit}{}@*)    def _rate_limit(self, side, pose, dt, reasons):
        # Function: cap translation and rotation jumps after all task edits.
        pose[:3, 3] = _limit_translation_step(previous[:3, 3], pose[:3, 3], max_translation_speed * dt)
        pose[:3, :3] = _limit_rotation_step(previous[:3, :3], pose[:3, :3], max_angular_speed * dt)
        return pose


(*@\hypertarget{code:wine-task-defaults}{}@*)def _apply_task_defaults(args, base_argv, config):
    # Function: when the guardrail wrapper is enabled, tighten the raw
    # retargeter and CRAFT smoothing defaults unless the user already passed
    # those base-script flags explicitly.
    defaults = {
        "translation_alpha": 0.65,
        "rotation_alpha": 0.75,
        "max_target_translation_speed": 0.36,
        "max_target_angular_speed": 2.80,
        "max_input_jump": 0.25,
        "max_input_rotation_jump": 1.25,
        "max_arm_joint_step": 0.045,
        "max_gripper_speed": 0.55,
        "max_step_raw": 56,
        "thumb_max_step_raw": 130,
        "max_velocity_raw": 950,
        "side_max_velocity_raw": 360,
    }
    apply_defaults_that_were_not_overridden(args, base_argv, defaults)


def _patch_retargeter(base, config_holder):
    # Function: replace QuestToYamRetargeter with a subclass whose update()
    # filters raw left/right arm poses before the raw loop commands them.
    class GuardedQuestToYamRetargeter(base.QuestToYamRetargeter):
        def calibrate(self, frame, left_eef_pose, right_eef_pose, now=None):
            ok = super().calibrate(frame, left_eef_pose, right_eef_pose, now=now)
            if ok:
                self._wine_guardrail.reset(left_eef_pose, right_eef_pose)
            return ok
        def update(self, frame, dt, now=None):
            output = super().update(frame, dt, now=now)
            return self._wine_guardrail.filter_output(output, dt)
    base.QuestToYamRetargeter = GuardedQuestToYamRetargeter


(*@\hypertarget{code:wine-guard-craft-targets}{}@*)def guard_craft_targets(targets, config):
    # Function: cap CRAFT side-splay, thumb bend, and finger bend so the glass
    # grasp stays gentle and does not over-close around the glass.
    for motor_id, raw in targets.items():
        if motor_id in SIDE_MOTOR_IDS:
            guarded[motor_id] = _limit_side_motor(motor_id, raw, craft_side_max_fraction, ...)
        elif motor_id in thumb_motor_ids:
            guarded[motor_id] = _limit_bend_motor(motor_id, raw, craft_thumb_max_fraction, ...)
        else:
            guarded[motor_id] = _limit_bend_motor(motor_id, raw, craft_grip_max_fraction, ...)
    return guarded


def _patch_craft_retarget(base, config_holder):
    # Function: wrap raw retarget_craft so CRAFT motor targets are filtered
    # before write_raw/async submit sends them to the hand.
    targets, signals, last_seen = original_retarget_craft(...)
    guarded_targets = guard_craft_targets(targets, config)
    return guarded_targets, signals, last_seen
\end{lstlisting}
\end{nolinenumbers}

\end{document}